\documentclass[UTF8]{article} 
\usepackage{iclr2023_conference,times}
\usepackage{microtype}
\usepackage{amsmath}
\usepackage{amssymb}
\usepackage{booktabs}
\usepackage{colortbl}
\usepackage{CJKutf8}
\usepackage[utf8]{inputenc}
\definecolor{lightgray}{rgb}{0.9,0.9,0.9}
\usepackage{caption}
\usepackage{multicol}
\usepackage{wrapfig}
\usepackage{subcaption}
\usepackage{xcolor}
\usepackage{graphicx}
\usepackage{setspace}
\usepackage{hyperref}
\usepackage{url}
\usepackage{multirow}
\usepackage{colortbl}
\usepackage{tabularx}
\usepackage{blindtext}
\usepackage{pgfplots}
\pgfplotsset{compat=1.18} 
\usepackage{tikz}
\usetikzlibrary{er,positioning,bayesnet}
\usepackage[inline]{enumitem}
\usepackage{makecell}
\usepackage{tipa}
\usepackage{siunitx}
\usepackage{tocloft}
\usepackage{algorithm} 
\usepackage{algpseudocode}  
\usepackage{listings}
\usepackage[raster,skins]{tcolorbox} 
\usepackage{xltabular}
\usepackage{hyperref}
\usepackage{amsmath}   
\usepackage{bm}        
\usepackage[framemethod=tikz]{mdframed}
\usepackage{todonotes}
\usepackage{fontawesome5}
\surroundwithmdframed[
  hidealllines=true,
  innerleftmargin=0pt,
  innertopmargin=0pt,
  innerbottommargin=0pt]{lstlisting}
\lstnewenvironment{response}[1][] 
 {\lstset{
 columns=fullflexible,
  breakautoindent=false, breakindent=0pt, breaklines, linewidth=8cm, #1}}
 {}
\tcbuselibrary{breakable}

\newcommand{\method}{\texttt{Yume}\xspace}
\newcommand{\sekai}{\texttt{Sekai}\xspace}

\newlength\savewidth

\definecolor{url_color}{RGB}{113, 187, 179}

\hypersetup{
    colorlinks=true,
    linkcolor=black,
    urlcolor=url_color,
}

\title{
\method: An Interactive World Generation Model}
\author{
    \hspace{-0.3em}\textbf{Xiaofeng Mao$^{1,2}$, Shaoheng Lin$^{1}$, Zhen Li$^{1}$, Chuanhao Li$^{1}$, Wenshuo Peng$^{1}$}, \\
    \textbf{Tong He$^{1}$, Jiangmiao Pang$^{1}$, Mingmin Chi$^{2\ddagger}$, Yu Qiao$^{1}$, Kaipeng Zhang$^{1,3\dagger\ddagger}$} \\
$^1$\normalfont{Shanghai AI Laboratory}~~~$^2$\normalfont{Fudan University}~~~$^3$\normalfont{Shanghai Innovation Institute}\\
\hspace{0.1em} ~\faGithub~ \hspace{0.4em} Github:\url{https://github.com/stdstu12/YUME}\\
\hspace{0.35em}~\faLaugh~ \hspace{0.1em} \hspace{0.1em} Huggingface:\url{https://huggingface.co/stdstu123/Yume-I2V-540P}\\
\hspace{0.25em}~\faYoutube~  \hspace{0.4em} Project Page: \url{https://stdstu12.github.io/YUME-Project}\\
\hspace{0.35em}~\faDatabase~ \hspace{0.55em} Data: \url{https://github.com/Lixsp11/sekai-codebase}
}
\iclrfinalcopy
\begin{document}

\maketitle

\if TT\insert\footins{\noindent\footnotesize\\
    This work was done during Xiaofeng's internship at Shanghai AI Laboratory.\\
    $^{\dagger}$~Project Leader\\
    $^{\ddagger}$~Corresponding Author \\
}\fi

\begin{abstract}
\method aims to use images, text, or videos to create an interactive, realistic, and dynamic world, which allows exploration and control using peripheral devices or neural signals. In this report, we present a preview version of \method, which creates a dynamic world from an input image and allows exploration of the world using keyboard actions. To achieve this high-fidelity and interactive video world generation, we introduce a well-designed framework, which consists of four main components, including camera motion quantization, video generation architecture, advanced sampler, and model acceleration. First, we quantize camera motions for stable training and user-friendly interaction using keyboard inputs. Then, we introduce the Masked Video Diffusion Transformer~(MVDT) with a memory module for infinite video generation in an autoregressive manner. After that, training-free Anti-Artifact Mechanism (AAM) and Time Travel Sampling based on Stochastic Differential Equations (TTS-SDE) are introduced to the sampler for better visual quality and more precise control. Moreover, we investigate model acceleration by synergistic optimization of adversarial distillation and caching mechanisms. We use the high-quality world exploration dataset \sekai to train \method, and it achieves remarkable results in diverse scenes and applications. \textbf{All data, codebase, and model weights are available on https://github.com/stdstu12/YUME.} \method will update monthly to achieve its original goal.

\textbf{\color{red}We are looking for collaboration and self-motivated interns interested in interactive world generation. Contact: zhangkaipeng@pjlab.org.cn}
\end{abstract}

\vspace{-5mm}
\begin{figure}[h]  
\centering
\includegraphics[width=0.9\textwidth]{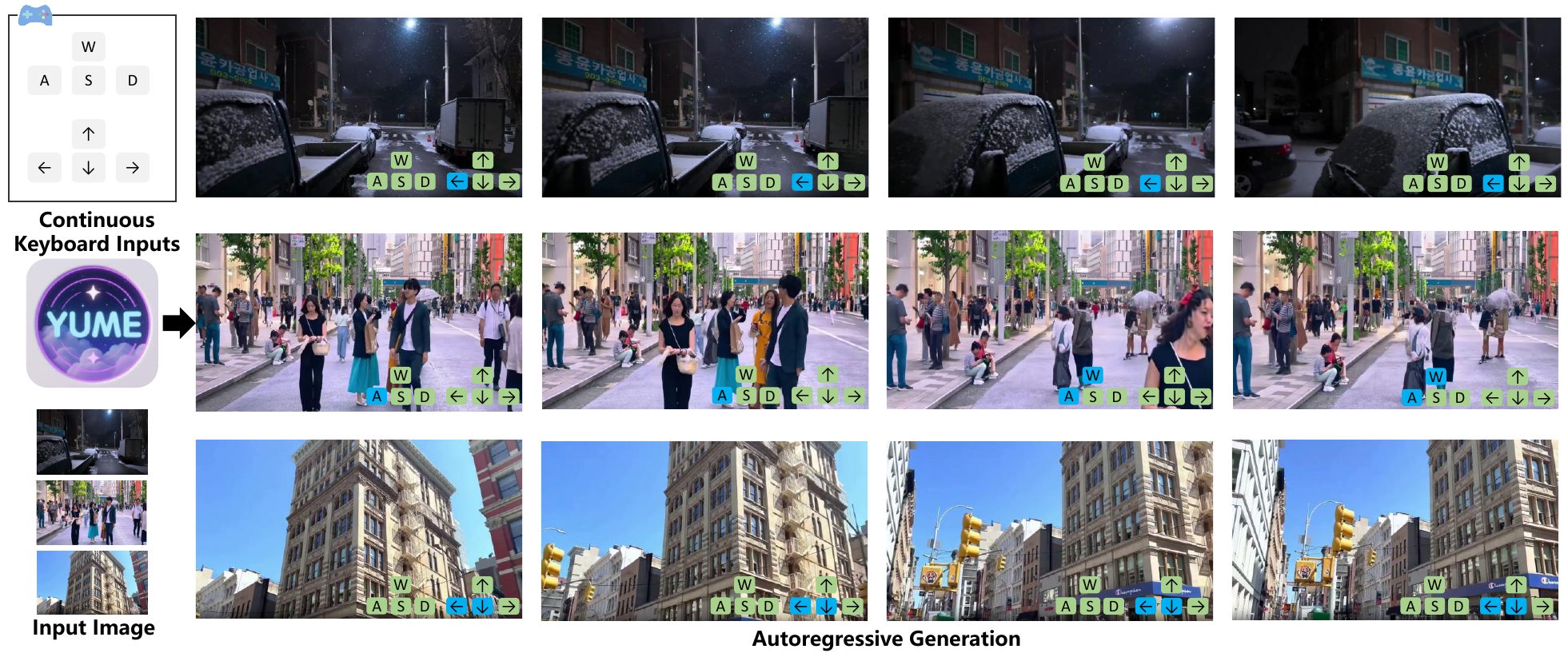} 
\captionof{figure}
{We introduce \method, a streaming interactive world generation model, which allows using continuous keyboard inputs to explore a dynamic world created by an input image.}
\label{banner}
\end{figure}

\clearpage

\newpage

\newcommand{\customsize}{\fontsize{7.7}{9}\selectfont}

\begingroup
\footnotesize
\begin{spacing}{0.88}
\tableofcontents
\end{spacing}
\endgroup

\hypersetup{linkcolor=url_color}

\newpage


\section{Introduction}
\label{sec:introduction}


\textit{\textbf{Past or future, just let them be} \ --- Link Click}

\method aims to create an interactive, realistic, and dynamic world through the input of text, images, or videos. And we can explore and control the world using peripheral devices or neural signals~\cite{zhou2025neural}. A practical application is to enter a world through a photo, like time travel, regardless of its location, scenes, and time, while you can interact with everything and change everything following your mind. We can compensate for our real-world regrets or realize our wishes through the virtual world created by \method. In this paper, we present a preview version of \method, which is an interactive world generation model that allows the use of keyboard inputs to explore a dynamic world created by an input image. We have released all data, code, and model weights on \url{https://github.com/stdstu12/YUME}. \method will update monthly to achieve its original goal.

Video diffusion models~\cite{Singer2023makeavideo,BarTal2024lumiere,Blattmann2023stablevideo,Ma2024hunyuanvideo,Nagrath2024mochi,Singer2023makeavideo,StepVideo2025technical,Chen2025skyreelsv2}, which have shown remarkable capabilities in synthesizing high-fidelity and temporally coherent visual content~\cite{Ho2022imagenvideo, Blattmann2023stablevideo}, present a promising avenue to realize such a sophisticated interactive world generation task. Recently, the automated generation of vast, interactive, and persistent virtual worlds~\cite{zhang2025matrixgame,agarwal2025cosmos} has advanced rapidly, driven by progress in generative models and the growing demand for immersive experiences in domains such as world simulation, interactive entertainment~\cite{li2025hunyuan}, and virtual embodiment~\cite{xiao2025worldmem, zhang2025matrixgame}.

The motion of the camera is an important control signal for interactive video generation of world exploration. However, existing video diffusion methods face significant challenges when applied to producing interactive (continuous camera motion controlled) and realistic videos, especially in urban scenarios. 
First, existing approaches~\cite{zhang2025matrixgame,xiao2025worldmem} focus mainly on synthetic or static scenarios, in terms of methods and data. 
The domain gap between them and the real world limits their generalizability. 
In addition, these methods are mostly based on absolute camera motions, which require precise annotation and extra-learner modules that increase the training and architecture design difficulty.
In addition, urban environments present unique complexities characterized by diverse architectural styles, dynamic objects, and intricate details. Existing approaches demonstrate limited adaptability to such complexity and struggle to maintain consistent realism across varied scenes. Common visual artifacts, including flickering, unnatural textures, and geometric distortions, significantly degrade the perceptual quality and disrupt immersive experiences.

To address these limitations, we introduce \method for the generation of autoregressive interactive video through an input image with discrete keyboard control.
\method is designed to offer more intuitive and stable camera control via keyboard input while improving the visual quality and realism of complex scene generation. Implements systematic optimizations on four key dimensions: camera motion control, model architecture, sampler, and model acceleration.

First, we proposed \textbf{Quantized Camera Motion~(QCM)} control module. \method quantizes camera trajectories into translational movements (forward, backward, left, right) and rotational motions (turn right, turn left, tilt up, tilt down), which can be flexibly combined and transferred by keyboard input. 
The QCM is produced by changes in relative camera poses during training and naturally embeds temporal context and spatial relationships into the control signal. QCM are parsed into textual conditions without introducing new learnable modules for pretrained I2V foundation models. 

Second, as noted in~\cite{zhang2025matrixgame}, text-based control frequently leads to unnatural outputs. To address this limitation, we investigated an alternative approach using \textbf{Masked Video Diffusion Transformers (MVDT)}, inspired by prior work~\cite{gao2023masked}. Moreover, we achieved interactive video generation with theoretically infinite duration via a chunk-based autoregressive generation framework with a modified FramePack memory module~\cite{zhang2025packing}. 

Third, to improve the visual quality of sophisticated real-world scenes, we introduce a training-free \textbf{Anti-Artifact Mechanism~(AAM)} module. Specifically, it refines the high-frequency components of the latent representation at each diffusion step. This targeted refinement improves fine-grained details, smooths out inconsistencies, and substantially reduces visual artifacts, leading to an overall improvement in visual quality without requiring any additional model training or specialized datasets. However, it does not perform well on autogressive long video generation due to a lack of V2V foundation models, and we will add it to our long video generation in the next version of \method.

Furthermore, we propose a novel sampling method based on the \textbf{Time-Travel Stochastic Differential Equation~(TTS-SDE)} framework. Inspired by DDNM \cite{wang2022zero} and OSV \cite{mao2025osv}, our approach leverages information from later denoising stages to guide the earlier denoising process, while incorporating stochastic differential equations to enhance sampling randomness, thereby improving textual controllability.

Finally, we investigate multiple acceleration techniques for diffusion-based video generation and \textbf{jointly optimize step distillation and cache mechanisms}. It significantly enhances sampling efficiency without compromising visual fidelity or temporal coherence.

    

\method offers a significant step towards generating high-quality, dynamic, and interactive infinite video generation, particularly for complex real-world scenes exploration.

\section{Related Works}
\label{sec:RelatedWorks}

\subsection{Video Diffusion Models}
Diffusion models~\cite{SohlDickstein2015deep, Ho2020denoising}, initially transformative for image synthesis, have rapidly become the cornerstone of video generation. The application of these models in a compressed latent space, pioneered by Latent Diffusion Models (LDMs), was a key enabler for efficient video synthesis, leading to works like Video LDM (Align your Latents)~\cite{Blattmann2023align} which extended this paradigm to achieve high-resolution video generation by integrating temporal awareness.

Early breakthroughs in text-to-video generation, such as Imagen Video~\cite{Ho2022imagenvideo} and Make-A-Video~\cite{Singer2023makeavideo}, quickly demonstrated the potential for creating dynamic scenes from textual descriptions. The field has since witnessed significant advancements in model scale, architectural design, and training strategies. Large-scale models like Google's Lumiere~\cite{BarTal2024lumiere}, featuring a Space-Time U-Net, and OpenAI's Sora~\cite{Brooks2024sora}, which employs a diffusion transformer architecture, have significantly pushed the boundaries of generating long, coherent, and high-fidelity video content.
Concurrently, the open-source ecosystem has flourished, with contributions such as Stable Video Diffusion~\cite{Blattmann2023stablevideo} offering robust and accessible baselines. Recent notable open-source efforts include HunyuanVideo~\cite{Ma2024hunyuanvideo}, which provides a systematic framework for very large video models, MoChi-Diffusion-XL~\cite{Nagrath2024mochi}, focusing on efficient high-resolution video synthesis, Step-Video-T2V~\cite{StepVideo2025technical}, a large-parameter foundation model, and SkyReels-V2~\cite{Chen2025skyreelsv2}, which aims for generating extended film-like content. Based on these advanced foundation models, we built \method for realistic dynamic world exploration.

\subsection{Camera Control in Video Generation}
\label{sec:related_cam_control}
Precise and flexible camera control is crucial for creating compelling and customizable video content, allowing for the emulation of cinematic effects and a more interactive user experience. Early video generation models often lacked explicit mechanisms for detailed camera path guidance, with camera movements emerging implicitly from textual prompts or initial frames. Subsequent research has focused on more direct methods for specifying and integrating camera motion into the generation process.

A significant body of work has emerged to explicitly condition video diffusion models on camera parameters. For instance, MotionCtrl~\cite{wang2023motionctrl} introduced a unified controller to manage both camera and object motion, where camera motion is determined by a sequence of camera poses that are temporally fused into the video generation model. Following this, Direct-a-Video~\cite{yang2024directavideo} enabled decoupled control of camera pan/zoom and object motion, utilizing temporal cross-attention layers for interpreting quantitative camera movement parameters, trained via a self-supervised approach. CameraCtrl~\cite{he2024cameractrl} proposed a plug-and-play module to integrate accurate camera pose control into existing video diffusion models, exploring effective trajectory parameterizations. Further advancing fine-grained control, CameraCtrl~II~\cite{zhang2024cameractrlii} focused on dynamic scene exploration by allowing iterative specification of camera trajectories for generating coherent, extended video sequences. More recently, training-free approaches have also been investigated; for example, CamTrol (Training-free Camera Control)~\cite{geng2024trainingfree} leverages 3D point cloud modeling from a single image and manipulates latent noise priors to guide camera perspective without requiring model fine-tuning or camera-annotated datasets. These methods typically rely on explicit sequences of absolute camera pose matrices or derived parameters to define the camera trajectory, aiming to improve the stability and precision of the generated viewpoint transformations. In contrast to existing approaches that rely on explicit camera trajectories~(requiring fine-grained motion parameter adjustments), we innovatively propose a Quantized Camera Motion (QCM) mechanism, which achieves intuitive keyboard-based control by discretizing the camera pose space.

\subsection{Navigatable World Generation}
\label{sec:related_nav_world}
The generation of expansive, interactive, and temporally coherent virtual worlds is a long-standing ambition in artificial intelligence, with significant implications for gaming, simulation, and robotics. Early efforts in this domain often involved learning world models~\cite{ha2018worldmodels} that capture environment dynamics to enable agents to plan or learn behaviors within a learned latent space, as demonstrated by subsequent lines of work like the Dreamer series~\cite{hafner2023dreamerv3}. These models emphasized understanding and predicting future states based on actions, laying the groundwork for more explicit world generation.

More recent approaches have focused directly on generating interactive environments and controllable long-duration video sequences. For instance, Genie~\cite{bruce2024genie} introduced a foundation model capable of generating an endless variety of action-controllable 2D worlds from image prompts, trained on unlabeled internet videos. Efforts to generate navigable driving scenes, such as GAIA-1~\cite{Wayve2023gaia1} by Wayve, have demonstrated the potential for creating realistic, controllable driving experiences from text and action inputs. Similarly, projects like SIMA~\cite{GoogleDeepMind2024sima} aim to develop generalist AI agents that can understand and interact within diverse 3D virtual settings based on natural language instructions.
The challenge of maintaining long-term consistency in generated worlds, especially for extended exploration, is actively being addressed. For example, StreamingT2V~\cite{henschel2024streamingt2v} proposed methods for consistent and extendable long video generation. Very recent works are pushing the envelope on creating interactive foundation models for worlds and ensuring their coherence. Matrix-Game~\cite{zhang2025matrixgame} presents an interactive world foundation model aimed at controllable game world generation, emphasizing fine-grained action control. Concurrently, WORLDMEM~\cite{xiao2025worldmem} introduces a framework to enhance long-term consistency in world simulation by employing a memory bank and attention mechanisms to accurately reconstruct previously observed scenes, even with significant viewpoint or temporal changes, and to model dynamic world evolution. These advancements are critical for generating the kind of scene-level, interactive, and navigable experiences that are the focus of current research. However, these methods primarily rely on game-based scenarios and actions (e.g., Matrix-Game ~\cite{zhang2025matrixgame} utilizes the Minecraft dataset), which remain relatively simplistic compared to real-world environments. Furthermore, collecting motion data from real-world video datasets is significantly more challenging than collecting it in game settings.

\subsection{Mitigating Generation Artifacts}
\label{sec:related_artifacts}
Despite the remarkable progress of diffusion models in generating realistic images and videos, the synthesized results can still suffer from various artifacts, such as unnatural textures, flickering, or semantic inconsistencies, particularly in complex scenes or long video sequences. Efforts to mitigate these issues can be broadly categorized into training-based and training-free approaches.

Training-based methods often involve architectural modifications or specialized fine-tuning strategies. For instance, some works focus on improving the autoencoder stage in latent diffusion models to better capture high-frequency details and reduce reconstruction errors that can propagate into the generation process, as explored in models like LTX-Video~\cite{hacohen2025ltxvideo}, which tasks the VAE decoder with a final denoising step. Others employ parameter-efficient fine-tuning (PEFT) techniques with objectives specifically designed to enhance visual quality and temporal consistency, as seen in frameworks like DAPE~\cite{xia2025dape} for video editing. Diffusion models themselves have also been adapted as post-processing tools to enhance the quality of already compressed or generated videos that may contain artifacts~\cite{liu2023diffusioncompressionvideo}.

While training-free methods typically operate during the inference or sampling stage, which often involves manipulations of the latent space, guidance of the denoising process, or modifications to sampling strategies. For video generation, FreqPrior~\cite{yuan2025freqprior} introduces a novel noise initialization strategy by refining noise in the frequency domain to improve detail and motion dynamics. Enhance-A-Video~\cite{luo2025enhanceavideo} proposes a training-free module to enhance the coherence and visual quality of videos from DiT-based models by adjusting temporal attention distributions during sampling. These methods aim to improve perceptual quality and reduce artifacts without altering the underlying model weights, offering flexible solutions for enhancing generated content. The exploration of latent space refinement, especially concerning frequency components and detail enhancement during the denoising process, remains an active area of research for improving the visual fidelity of diffusion-based generation.

\subsection{Video Diffusion Acceleration}
Numerous distillation methodologies~\cite{wang2023videolcm,song2023consistency,wang2024animatelcm,wang2024animatelcm-1,wang2025phased,wang2024rectified,mao2024osv,kim2023consistency} have been proposed to reduce computational costs in video generation. For instance, some approaches implement joint consistency distillation and adversarial training~\cite{wang2025phased}, accelerating diffusion models through phased consistency models integrated with GANs. Concurrently, Mao \emph{et al.}~\cite{mao2024osv} enhanced the discriminator architecture for adversarial distillation. Given the feature similarity across timesteps in DiT architectures, cache mechanisms have been developed for diffusion frameworks, such as ToCa~\cite{zou2024accelerating}'s dynamic feature storage guided by token sensitivity and error propagation analysis, which accelerates diffusion transformers through adaptive caching strategies and layer-specific retention techniques. Similarly, AdaCache~\cite{kahatapitiya2024adaptive} improves diffusion transformer inference without retraining via dynamically adjusted caching policies and motion-aware resource allocation during denoising steps, and TeaCache~\cite{liu2024timestep} accelerate sampling via estimating fluctuating differences among model outputs across timesteps. We introduced a co-optimization strategy that integrates step distillation with cache acceleration to further boost sampling efficiency of interactive video generation.

\section{Preliminaries}
\subsection{Rectified Flow. }
Rectified Flow \cite{liu2022flow} is a technique that facilitates ordinary differential equation (ODE)-based training by minimizing the transport cost between marginal distributions $\pi_0$ and $\pi_1$, i.e., $ \mathbb{E}[c(x_{1}-x_{0})]$. Here, $x_{1} = \mathrm{Law}(\pi_1)$, $x_{0} = \mathrm{Law}(\pi_0)$, and $c : \mathbb{R}^{d} \longrightarrow \mathbb{R}$ denote a cost function. Given the computational complexity of Optimal Transport (OT) \cite{villani2009optimal}, Rectified Flow provides a simple yet effective approach to generate a new coupling from a preexisting one. This new coupling can be optimized using Stochastic Gradient Descent (SGD), an optimization method extensively employed in deep learning:
\begin{equation}
\label{eq:1}
\displaystyle
\theta^{*}= \underset{\theta}{\arg \min} \mathbb{E}_{t\sim U[0,1]}
\mathbb{E}_{x_{0},x_{1}\sim \pi_0,\pi_1}
\left[\text{MSE}\left(x_{1}-x_{0}, v_{\theta}(x_{t}, t)\right)\right],
\end{equation}
where the term $\text{MSE}$ denotes Mean Squared Error. The parameter $t$ is in the range of [0, 1], we select $x_{t} = tx_{1} + (1-t)x_{0}$ to ensure that $\forall v_{\theta}(x_{t}, t)$ matches the identical target velocity $x_{1}-x_{0}$. Upon completion of the training phase, sampling can be performed through a definite integral, specifically $x_{1}=x_{0}+\int_{0}^{1} v_{\theta}(x_\tau,\tau)\mathrm{d}\tau$. In practical scenarios, the aforementioned continuous system is typically approximated in discrete time using the Euler method (or its variants): $x_{t_{n-1} } = x_{t_n}+(t_{n-1}-t_{n}) v_{\theta}(x_{t_n}, t_n)$, where $ t_0 < t_1 < \ldots < t_{N-1} $ is a set of predefined time steps.

In the video domain, $x_0$ represents the pixel space of the videos, while $c$ denotes the conditioning inputs (such as text and image conditions for controlled generation). Following recent advancements in Rectified Flow that employ VAEs~\cite{rombach2022high} for latent space compression during both training and inference to reduce computational costs, we formulate the training objective as:
\begin{equation}
\label{eq:2}
\theta^{*}= \underset{\theta}{\arg \min} \mathbb{E}_{t\sim U[0,1]}
\mathbb{E}_{z_{0},z_{1}\sim \pi_0,\pi_1}
\left[|| z_{1}-z_{0} - v_{\theta}(z_{t}, c, t)||^2_2\right],
\end{equation}
where $z=\text{VAE\_Encoder}(x)$ and $x=\text{VAE\_Decoder}(z)$ perform latent space projection and reconstruction respectively. During inference, the sampling process follows: $z_{t_{n-1}} = z_{t_n} + (t_{n-1}-t_{n}) v_{\theta}(z_{t_n}, c, t_n).$

\subsection{Wan Architecture}\label{wan}
The foundational architecture employs an identical design to Wan~\cite{wan2025wan}, utilizing its spatio-temporal VAE encoder and denoising DiT model. The VAE encoder compresses input video sequences into latent representations of dimensionality [1 + T/4, H/8, W/8] with expanded channel depth C=16. The denoising DiT backbone incorporates: 1) Patchify module utilizing 3D convolution (kernel=(1,2,2) to downsample spatial resolution while expanding channels into transformer tokens; 2) CLIP~\cite{radford2021learning} for image encoder and umT5~\cite{chung2023unimax} for text encoder; 3) Transformer blocks that concurrently process modality-specific features, where cross-attention mechanisms fuse the video tokens (queries) with image/text embeddings (keys/values). This configuration maintains spatio-temporal coherence while ensuring computational efficiency throughout the encoding-to-denosing pipeline.

\section{Data Processing}
\label{sec:Data}

\subsection{Dataset}
We use the Sekai-Real-HQ, a subset of \sekai~\cite{li2025sekai}, as the training dataset. It consists of large-scale walking video clips with corresponding high-quality annotations of camera trajectory and semantic labels. In this section, we briefly introduce the dataset and more details could be found in the paper of \sekai.

\textbf{Video Collection}
We manually collect high-quality video URLs from popular YouTubers and extend them by searching additional videos using related keywords (e.g., walk, drone, HDR, and 4K). In total, we collect 10471 hours of walking videos (with stereo audio) and 628 hours of drone (FPV or UAV) videos. All videos were released over the past three years, with a 30-minute to 12-hour duration. They are at least 1080P with 30 to 60 FPS. We download the 1080P version with the highest Mbps for further video processing and annotation. Due to network issues and some videos are broken, there are 8409 hours of walking videos and 214 hours of drone videos after downloading. 

\textbf{Video Preprocessing }
For YouTube videos, we trim two minutes from the start and end of each original video to remove the opening and ending. Then we do the five preprocessing steps, including shot boundary detection,  clip extraction and transcoding, luminance filtering, quality filtering, camera trajectory filtering, to obtain 6620 hours of video clips as Sekai-Real. 

\textbf{Video Annotation}
We annotate video clips using multiple tools, large vision-language models, and meta inforation from YouTube. The annotations including location, multiple categories, caption, and camera trajectories.

\textbf{Video Sampling}
We sample the best-of-the-best video clips considering content diversity, location diversity, category diversity, and camera trajectory diversity. Finally, we obtain 400 hours of video clips as Sekai-Real-HQ.

\subsection{Camera Motion Quantization}
\label{sec:Method_Quantized_Camera_Motion}
Though low-quality camera trajectories from Sekai-Real-HQ are already filtered, the trajectories estimated by MegaSaM~\cite{li2025megasam} are inevitably imprecise enough. In addition, raw camera trajectories with large variances are difficult to follow and require extra learnable modules. Consequently, we developed a trajectory quantization method to transfer continuous camera poses to discrete actions. This approach inherently facilitates filtering excessively jittery trajectories, balances trajectory distributions, and mitigates model training difficulty.


Specifically, existing approaches for camera control in video generation often rely on providing a dense sequence of per-frame camera-to-world (c2w) transformation matrices~\cite{wang2023motionctrl, he2024cameractrl}. While offering explicit control, this representation can be overly granular, potentially leading to less stable or unintuitive camera motion, and may not effectively capture the inherent temporal coherence of continuous camera movements.

To address these limitations and foster more robust and intuitive navigation, \method introduces a quantized camera motion representation. Our core idea is to define a discrete set of predefined, holistic camera motion, $\mathbb{A}_{\text{set}}$. Each motion $A_j \in \mathbb{A}_{\text{set}}$ (e.g., ``move-forward'', ``tilts up'', ``tilts down''. See Supplementary Materials for these motion.) corresponds to a canonical relative transformation matrix, $T_{\text{canonical},j}$, representing a typical navigational maneuver. Instead of directly using the transformation matrices, we process the input sequence of c2w transformation matrices. For each segment of camera movement (defined by a pair of consecutive transformation matrices, potentially after downsampling the trajectory), we construct the reference coordinate system using the transformation matrix at the current segment, and calculate the relative transformation matrix $T_{\text{rel,actual}}$ of the next segment. We then select the motion $A_j^*$ from our predefined set $\mathbb{A}_{\text{set}}$ whose canonical transformation matrix $T_{\text{canonical},j}$ is closest to $T_{\text{rel,actual}}$, as outlined in Algorithm~\ref{alg:quantized_camera_action_selection}. This matching process effectively quantizes the continuous camera trajectory into a sequence of semantically meaningful motion, inherently integrating temporal context from the relative pose changes.

In Algorithm~\ref{alg:quantized_camera_action_selection}, the $\text{Distance}(T_1, T_2)$ function measures the dissimilarity between two transformation matrices, which can be a weighted combination of differences in their translational and rotational components.

\begin{figure*}[t]
    \centering
    \includegraphics[width=1.0\linewidth]{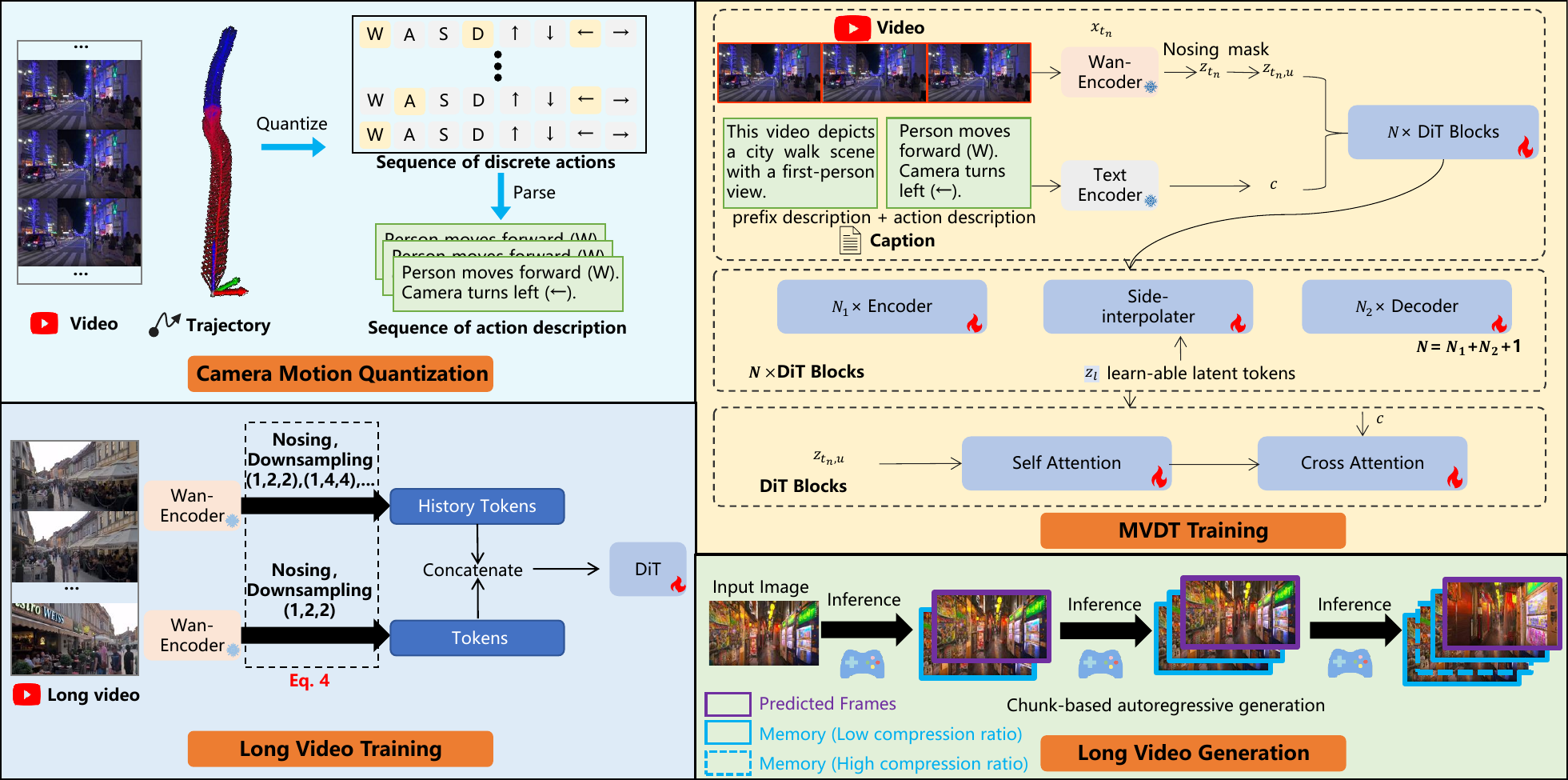}
    \caption{Four core components of \method: camera motion quantization, model architecture, long video training, and generation. We also make advanced sampler, please see Section~\ref{sampler}.}
    \label{fig:pipeline}
\end{figure*}

As shown in Figure\textcolor{red}{~\ref{fig:pipeline}}, the selected discrete motion $A^*$ are assigned to textual descriptions through a predefined dictionary (detailed in the Supplementary Materials). By injecting the action descriptions into the text condition, this approach achieves camera pose-controlled video generation without introducing additional learnable parameters to the pre-trained I2V model, leading to fast model converged and stable and precise camera control.

By quantifying camera motion for videos from Sekai-Real-HQ and extracting segments with consistent camera movements, we selected clips longer than 33 frames, ultimately obtaining 139019 video clips. The distribution of camera motions is provided in the supplementary materials.
\begin{algorithm}
\caption{Camera Motion Quantization}
\label{alg:quantized_camera_action_selection}
\begin{tabular}{rl}
\multicolumn{2}{l}{{\bf Require:} Sequence of camera-to-world matrices $\mathcal{C} = \{C_0, C_1, \ldots, C_{N-1}\}$,} \\
\multicolumn{2}{l}{\quad Predefined set of $K$ camera motion $\mathbb{A}_{\text{set}} = \{A^{(1)}, \ldots, A^{(K)}\}$,} \\
\multicolumn{2}{l}{\quad Corresponding canonical relative $SE(3)$ transformations} \\
\multicolumn{2}{l}{\quad $\{T_{\text{canonical}}^{(1)}, \ldots, T_{\text{canonical}}^{(K)}\}$.} \\
\multicolumn{2}{l}{{\bf Ensure:} Sequence of selected camera motion $\mathcal{A}^* = \{A^*_0, \ldots, A^*_{M-1}\}$} \\
\multicolumn{2}{l}{\quad (where $M$ depends on processing stride).} \\
1: & Initialize $\mathcal{A}^* \leftarrow \emptyset$ \\
2: & {\bf for} each relevant pair $(C_{\text{curr}}, C_{\text{next}})$ from $\mathcal{C}$ {\bf do} \\
3: & \quad $T_{\text{rel,actual}} \leftarrow C_{\text{curr}}^{-1} \cdot C_{\text{next}}$ \\
4: & \quad $A^*_{\text{current}} \leftarrow \arg\min\limits_{A^{(j)} \in \mathbb{A}_{\text{set}}} \text{Distance}(T_{\text{rel,actual}}, T_{\text{canonical}}^{(j)})$ \\
5: & \quad Append $A^*_{\text{current}}$ to $\mathcal{A}^*$ \\
6: & {\bf end for} \\
7: & {\bf return} $\mathcal{A}^*$ \\
\end{tabular}
\end{algorithm}

\section{Method}
\label{sec:Method}
\subsection{Overview}

In this section, we provide a comprehensive overview of the \method. The architecture builds upon Wan~\cite{wan2025wan}'s network design but introduces masked video diffusion transformers and employs a ReFlow-based training methodology. To enable long-video generation, we concatenate downsampled historical video clips (using Patchify) with generated segments during training and feed them into the DiT denoising model—a strategy replicated during inference by similarly downsampling and stitching multiple video segments. Moreover, we detail various sampler designs, including AAM and TTS-SDE, which yield distinct sampling effects. Then we introduced our camera motion control and the motion speed control. Finally, we discuss practical applications of \method, including world generalization and world editing.

\subsection{Model Architecture}
\subsubsection{Masked Video Diffusion Transformers}
\label{sec:Masked_Video_Diffusion_Transformers}
We employ the video diffusion model described in Section~\ref{wan} and introduce Masked Video Diffusion Transformers~(MVDT) for better visual quality. 

Existing video diffusion models often overlook the critical role of masked representation learning, leading to artifacts and structural inconsistencies in cross-frame generation. Drawing inspiration from proven masking strategies in ~\cite{gao2023masked,mao2024mdt}, we introduce MVDT to significantly enhance video generation quality. As shown in Figure\textcolor{red}{~\ref{fig:pipeline}}, the architecture employs an asymmetric network to process selectively masked input features through three core components: encoder, side-interpolator, and decoder.

\textbf{Masking Operation.} The feature transformation begins with stochastic masking of the input feature $z_{t_n} \in \mathbb{R}^{N \times d}$ (where $d$ denotes the channel dimension and N represents the token count), derived from noisy video token embeddings. Applying random masking ratio $\rho$ yields a reduced set of active features $z_{{t_n},u} \in \mathbb{R}^{d \times \hat{N}} (\hat{N}=(1-\rho)N)$ accompanied by a positional binary mask $MASK \in \mathbb{R}^N$. The masking ratio $\rho$ serves as a  hyperparameter. We set $\rho = 0.3$ in this paper.

This selective processing concentrates computational resources on visible tokens $z_{{t_n},u}$ while maintaining representational accuracy and substantially reducing memory/computational overhead.

\textbf{Encoder. }This stage utilizes a streamlined architecture that exclusively processes preserved tokens $z_{{t_n},u}$, mapping them to compact latent representations. By bypassing masked regions, it achieves a computational load reduction compared to conventional full-feature encoders. 

\textbf{Side-Interpolator. }Inspired by prior work~\cite{gao2023masked}, this component adopts an innovative dual-path architecture. During training, it combines learnable latent tokens $z_l$ with encoded features through $z_I=SA(\mathrm{Concat}[z_l, Encoder(z_{{t_n},u})])$, dynamically predicting masked content via self-attention mechanisms. The gated fusion operation $\hat{z}_{{t_n},u}=(1-MASK)\odot z_{{t_n},u}+MASK\odot z_I$ seamlessly integrates the original and synthesized features while preserving temporal coherence between video sequences.

\textbf{Decoder. }Finally, we process the interpolated features $\hat{z}_{{t_n},u}$ using the remaining DiT-blocks.

During training, it optimizes MVDT through end-to-end gradient propagation, while inference directly processes complete feature sets without interpolation. During model training, the diffusion module receives both $z_{t_n}$ and $z_{t_n,u}$. This strategy prevents the system from concentrating exclusively on masked-area regeneration while compromising essential diffusion learning.

\begin{figure*}[htp]
    \centering
    \includegraphics[width=1.0\linewidth]{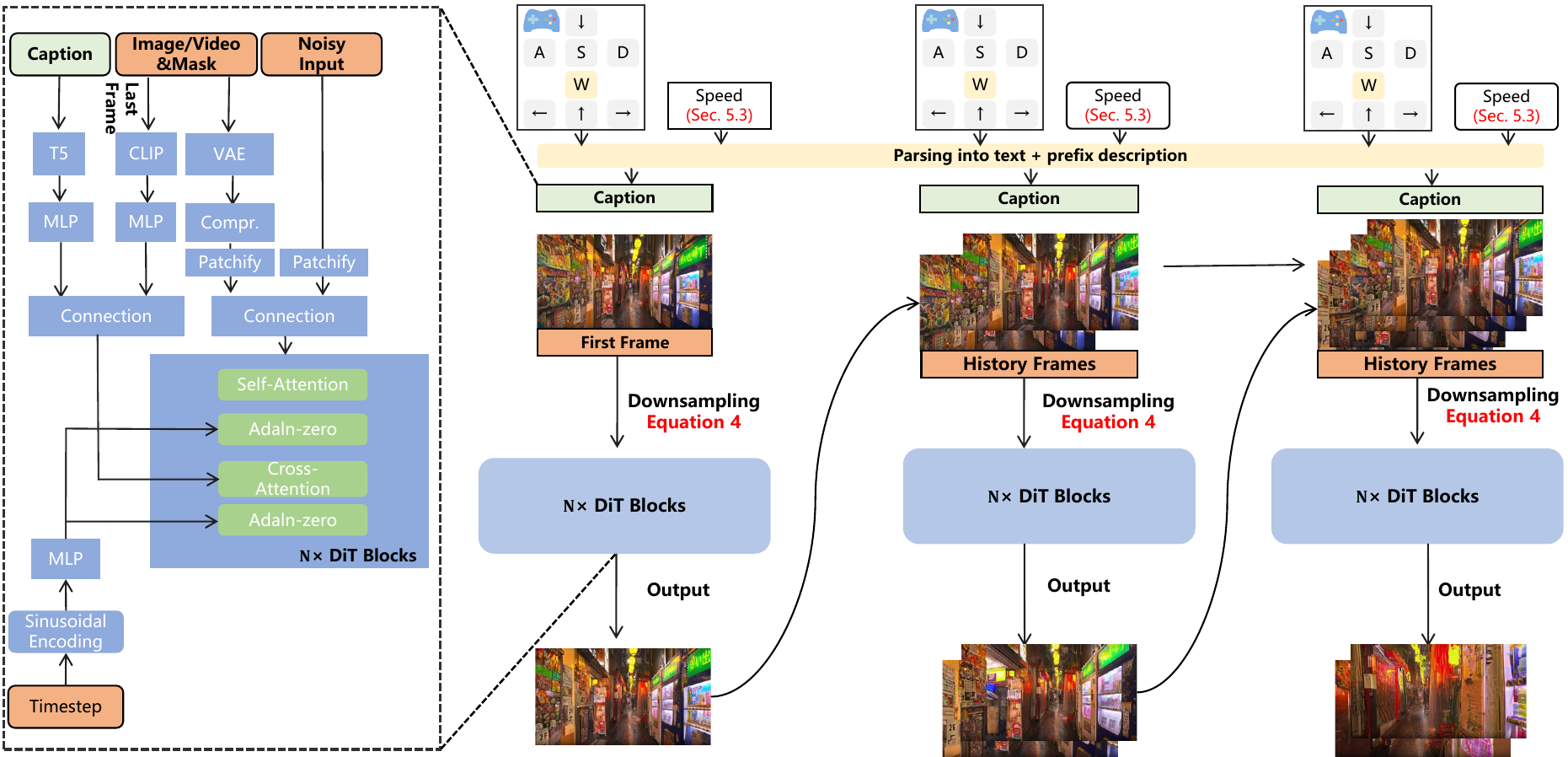}
    \caption{Long-form video generation method.}
    \label{fig:long}
\end{figure*}

\subsubsection{Image-to-Video~(I2V) and Video-to-Video~(V2V) Generation}
For image-to-video, we adopt the methodology established in Wan. In image-to-video synthesis, a conditioning image initializes video generation. Temporally zero-padded frames $x_c$ undergo compression via the Wan-VAE encoder to yield latent representations $z_c \in \mathbb{R}^{C \times t \times h \times w}$. 

A binary mask $M_c \in \{0,1\}^{1\times T\times h\times w}$ (with 1 denoting preserved regions and 0 indicating generated regions) is reshaped to dimensions $s \times t \times h \times w$ ($s$ denoting the Wan-VAE temporal stride). Channel-wise concatenation of $[z_t, z_c, M]$ feeds into Wan’s DiT backbone. An additional zero-initialized projection layer accommodates the increased channel dimensionality inherent to image-to-video tasks ($2c + s$ versus $c$ for text-to-video). For video-to-video generation, our approach diverges from the image-to-video methodology by extracting exclusively the final frame of the video condition as the image input for CLIP encoding. We concurrently adjust the masking tensor $M_c$ to ensure precise alignment between the conditional video sequence and generated output. This video-to-video synthesis lacks inherent memory retention capabilities.

\subsubsection{Long Video Generatetion}
Based on meticulous experimentation, we observe that autoregressive methods for long video generation frequently exhibit inter-frame discontinuity and lack temporal coherence. To address these limitations, we introduce a long-form video generation methodology, adopting a compression design analogous to FramePack~\cite{zhang2025packing} (architectural details illustrated in Figure~\textcolor{red}{\ref{fig:long}}). FramePack employs contextual compression of historically generated frames via downsampling of pretrained Patchify parameters. The baseline Patchify configuration, specified by hyperparameters (2, 4, 4), denotes sequential compression ratios of 2× temporally, 4× spatially in height, and 4× spatially in width. Higher compression ratios are applied as generated video length increases.

For conditional input to the diffusion model, our established compression scheme operates as follows:  
\begin{equation}
\begin{aligned}
&\text{Frame } t-1: & (1, 2, 2) \\
&\text{Frames } t-2 \text{ to } t-5: & (1, 4, 4) \\
&\text{Frames } t-6 \text{ to } t-22: & (1, 8, 8) \\
&\vdots & \vdots \\
\end{aligned}
\end{equation}
... and so forth, including scenarios with spatiotemporal compression.  

More high-resolution tokens generally lead to better performance. We finalized:  
\begin{equation}
\begin{aligned}
&\text{Frames } t-1 \text{ to } t-2: & (1, 2, 2) \\
&\text{Frames } t-3 \text{ to } t-6: & (1, 4, 4) \\
&\text{Frames } t-7 \text{ to } t-23: & (1, 8, 8) \\
&\vdots & \vdots \\
&\text{Initial frame:} & (1, 2, 2)
\end{aligned}
\end{equation}
This configuration optimally balances temporal and spatial compression while preserving high-resolution representation for recent frames and retaining critical features from initial input images.

We implement a stochastic sampling strategy for the training where, with 0.3 probability, we select historical frames from 33-400 frame videos to predict subsequent 33-frame sequences, and with 0.7 probability, we utilize 400-800 frame videos as historical context for 33-frame predictions. This sampling distribution optimizes the frame count allocation during training.

In our sampling methodology, letting ${z}_{t_n} \in \mathbb{R}^{C \times F \times H \times W}$ denote the latent representation at denoising step $t_n$, where $C$ indexes channels, $F$ frames, and $H \times W$ spatial dimensions. Given conditioning inputs ${c}$ (text embeddings), historical frames $ \mathbf{I}_{\text{input}}$, noise ${z}_{\text{noise}}$, and diffusion model $f_\theta$, our sampling process implements:
\begin{equation}
\begin{aligned}
z_{t_{n-1}} &= {z}_{\text{in}} + (t_{i-1}-t_i) \cdot v_\theta(\hat{z}_{\text{in}}, {c}, t_{n}) \\
\hat{z}_{\text{in}} &= ((1-t_{n})\mathbf{I}_{\text{in}} +t_n{z}_{\text{noise}}) \oplus z_{in}
\end{aligned}
\end{equation}
where $\oplus$ is frame-wise concatenation.

Our method supports both video-to-video (V2V) and image-to-video (I2V) generation. However, during V2V synthesis, we observe that when the conditioning video contains prolonged homogeneous motion, the model becomes overly reliant on the motion characteristics of the conditioning video (resulting in identical motion patterns between source and generated sequences), weakening the motion following capability. To address this conditional video dependency issue, we implement a training strategy where 30\% of iterations use static image conditions, specifically, by temporally tiling individual images 16 times to construct static video sequences, thereby improving the model's ability to generate diverse motion patterns.

\begin{algorithm}
\caption{Anti-Artifact by High Frequency Refinement}
\label{alg:anti_artifact_refinement}
\begin{tabular}{rl}
\multicolumn{2}{l}{{\bf Require:} Initial latent estimate $z_{\text{orig}}$ (from a full standard denoising pass),} \\
\multicolumn{2}{l}{\quad Diffusion model $v_\theta$, condition $c$, } \\
\multicolumn{2}{l}{\quad number of inference steps $N$, refinement steps $K_{refine} < N$, } \\
\multicolumn{2}{l}{\quad Timesteps $\{t_i\}_{i=0}^{N-1}$ from $T$ down to $1$,} \\
\multicolumn{2}{l}{\quad Low-pass filter operator $B$ (e.g., Gaussian blur).} \\
\\
1: & $z_{t_{N-1}}$ is initialized during the first denoising step. \\
2: & {\bf for} $i = N-1$ down to $0$ {\bf do} \\
3: & \quad $t_{i} \leftarrow \text{current timestep from schedule}$ \\
4: & \quad $z'_{\text{in}} \leftarrow z'_{t_{i}}$ \\\
5: & \quad {\bf if} $i \ge N - K_{refine}$ {\bf then} \\
6: & \quad\quad $z_{\text{orig},t_{\text{i}} } \leftarrow (1-t_{i})*z_{\text{orig}}+t_{\text{i}}*z_{t_{N-1}}$ \\
7: & \quad\quad $z_{\text{low\_from\_orig}} \leftarrow B(z_{\text{orig},  t_{i} })$ \\
8: & \quad\quad $z'_{\text{high\_current}} \leftarrow z'_{\text{in}} - B(z'_{\text{in}})$ \\
9: & \quad\quad $z'_{\text{in}} \leftarrow z_{\text{low\_from\_orig}} + z'_{\text{high\_current}}$ \\
10: & \quad {\bf end if} \\
11: & \quad $z'_{t_{i-1}} \leftarrow z'_{\text{in}}+(t_{i-1}-t_i)*v_\theta(z'_{\text{in}}, c, t_{i})$  \\
12: & {\bf end for} \\
13: & {\bf return} $z'_{t_0}$  \\
\end{tabular}
\end{algorithm}

\subsection{Sampler Design}
\label{sampler}
This section details our advanced sampler, which incorporates two key innovations to enhance the quality of the generated videos. We first introduce a Training-Free Anti-Artifact Mechanism (AAM) to eliminate visual artifacts and then present Time Travel Sampling based on SDE (TTS-SDE) to improve video sharpness and textual controllability.

\subsubsection{Training-Free Anti-Artifact Mechanism}
\label{sec:Training_Free_Anti_Artifacts}
While diffusion models excel at generating diverse content, complex scenes such as urban cityscapes can often exhibit visual artifacts, such as blurred details, unnatural textures, or flickering, which detract from realism. To address this without requiring additional training or model modification, \method incorporates a novel \textbf{Training-Free Anti-Artifact Mechanism, AAM}. The core idea is to enhance high-frequency details and overall visual quality by strategically refining the latent representation during a second stage of denoising pass, leveraging information from an initial standard generation. This approach draws inspiration from NVIDIA DLSS, which primarily employs neural networks to super-resolve low-resolution rendered frames in video games.
It is also inspired by DDNM~\cite{wang2022zero}, a training-free method achieving image deblurring, super-resolution, etc..

Our mechanism involves a two-stage process. Let $\mathbf{z}_{t_N} \in \mathbb{R}^{C \times F \times H \times W}$ denote the latent representation at denoising step $t_n$, where $C$ indexes channels, $F$ frames, and $H \times W$ spatial dimensions. First, a standard multi-step denoising process is performed, starting from random noise $z_{t_{N-1}}$ and conditioned on text and any other relevant input, to obtain an initial latent estimate, which we denote as $z_{\text{orig}}$. In the second stage, we employ the pre-trained diffusion model and remove motion control from the text conditioning, since $z_{\text{orig}}$ already incorporates trajectory characteristics. This $z_{\text{orig}}$ typically captures the overall structure and semantics of the scene well, but may contain the aforementioned artifacts or lack fine details.

Second, a refinement denoising pass is initiated, also starting from the sample of $z_{t_{N-1}}$. During the initial $K$ steps of this refinement pass (where $K$ is a small number, e.g., 5), we intervene in the denoising process before each model prediction. 

\begin{algorithm}
\caption{Time Travel Sampling based on SDE~(TTS-SDE)}
\label{alg:time_travel_sampling}
\begin{tabular}{rl}
\multicolumn{2}{l}{{\bf Input:} Initial conditioning ${c}$, model $V_\theta$,} \\
\multicolumn{2}{l}{\quad schedule $\{t_i\}_{i=0}^{N-1}$, and $t_{-1}=0$} \\
\multicolumn{2}{l}{\quad Travel parameters: interval $s=5$, depth $l=5$} \\
\multicolumn{2}{l}{{\bf Output:} Generated latent ${z}_0$} \\
\\
1: & Initialize noise ${z}_{t_{N-1}}$ \\
2: & {\bf for} $i = N-1 $ {down to} $1$ {\bf do} \\
3: & \quad $t_{i} \leftarrow \text{current timestep from schedule}$ \\
4: & \quad $z_{\text{in}} \leftarrow z^{\text{out}}_{t_{i}}$ \\
5: & \quad $z'_{t_{i-1}} \leftarrow z_{\text{in}}+(t_{i-1}-t_i)*v_\theta(z_{\text{in}}, c, t_{i})$ \\
6: & \quad $\hat{z}_{t_{0}} \leftarrow z_{\text{in}}+(0-t_i)*v_\theta(z_{\text{in}}, c, t_{i})$ \\
7: & \quad ${\beta_i}$ $\leftarrow$ -0.5$\eta^2 *\frac{-(z_{\text{in}}-\hat{z}_{t_{0}})}{t_{i}^2}$ \\
8: & \quad $\Delta\mathbf{z} \leftarrow \eta  \sqrt{|t_{i-1} - t_{i}|} \mathcal{N}(0,\mathbf{I})$ \\
9: & \quad $z_{t_{i+1}} \leftarrow z'_{t_{i-1}} + \beta_i (t_{i-1}-t_i) + \Delta\mathbf{z}$ \\
10: & \quad {\bf if} $t \equiv 0 \pmod{s}$  {\bf then} \\
11: & \quad\quad $k_{\text{max}} \leftarrow \max(t-l, 0)$ \\
12: & \quad\quad {\bf for} $k = i-1$ {\bf down to} $k_{\text{max}}$ {\bf do} \\
13: & \quad\quad\quad $t_{k} \leftarrow \text{current timestep from schedule}$ \\
14: & \quad\quad\quad $z'_{\text{in}} \leftarrow z_{t_{k+1}}$ \\
15: & \quad\quad\quad $z'_{t_{k-1}} \leftarrow z'_{\text{in}} +(t_{k-1}-t_k)*v_\theta(z'_{\text{in}}, c, t_{k})$ \\
16: & \quad\quad\quad $\hat{z}_{t_{0}} \leftarrow z'_{\text{in}}+(0-t_k)*v_\theta(z'_{\text{in}}, c, t_{k})$ \\
17: & \quad\quad\quad ${\beta_k}$ $\leftarrow$ -0.5$\eta^2 *\frac{-(z'_{\text{in}}-\hat{z}_{t_{0}})}{t_{k}^2}$ \\
18: &\quad\quad\quad $\Delta\mathbf{z} \leftarrow \eta  \sqrt{|t_{k-1} - t_{k}|} \mathcal{N}(0,\mathbf{I})$ \\
19: & \quad\quad\quad $z_{t_{k+1}} \leftarrow z'_{t_{k-1}} + \beta_k (t_{k-1}-t_k) + \Delta\mathbf{z}$ \\
20: & \quad\quad\quad $ {v}_k \leftarrow v_\theta(z'_{\text{in}}, c, t_{k})$ \\
21: & \quad\quad {\bf end for} \\
22: & \quad\quad\quad $ \hat{v} \leftarrow v_{k_{\text{max}}}$ \\
23: & \quad {\bf end if} \\
24: & \quad $z^{\text{out}}_{t_{i-1}} \leftarrow z_{\text{in}}+(t_{i-1}-t_i)*\hat{v}$ \\
25: & {\bf end for} \\
26: & return $z^{\text{out}}_{t_{0}}$ \\
\end{tabular}
\end{algorithm}

For the current latent $z'_{t_i}$ at timestep $t$~($t_i>=t_{N-K}$) in the refinement pass, we perform a detailed recombination as outlined in Algorithm~\ref{alg:anti_artifact_refinement}. We begin by diffusing (noising) the initial estimate $z_{\text{orig}}$ back to the current noise level corresponding to timestep $t_i$, resulting in $z_{\text{orig}, t_i}=(1-t_i)*z_{\text{orig}}+t_i*z_{t_{N-1}}$. From this noised estimate, we extract the low-frequency component using a low-pass filter (e.g., a blur operator $A$, $B(z) = A^{\text{Pinv}}Az$, the supplementary materials contain additional information about matrix $A$. Note that $A$ is not full-rank and $A^{\text{Pinv}}$ represents the pseudo-inverse of $A$), yielding $z_{\text{low\_from\_orig}} = B(z_{\text{orig}, t_i})$. This component represents the stable, coarse structure of the initial generation. Simultaneously, we extract the high-frequency component of the current refinement latent $z'_{t_i}$ by taking the difference between $z'_{t_i}$ and its low-frequency version, giving us 
\begin{equation}
z'_{\text{high\_current}} = z'_{t_i} - B(z'_{t_i}) 
\end{equation}
which can also be expressed as $(I - A^{\text{Pinv}}A)z'_t$, where $I$ is the identity matrix. The latent that will be fed into the diffusion model for the current denoising step is then recomposed by combining these complementary frequency components: 
\begin{equation}
z'_{t_i} \leftarrow z_{\text{low\_from\_orig}} + z'_{\text{high\_current}}
\end{equation}
This approach effectively merges the low-frequency information from the initial generation with the high-frequency information from the current refinement step. After this recomposition for the first $K$ steps, the diffusion model $v_\theta$ predicts the less noisy latent $z'_{t_{i-1}}$ as usual. For the remaining steps of the refinement pass (i.e., after the initial $K$ steps), the standard denoising procedure continues without this frequency-domain intervention.

This strategy allows the refinement pass to preserve the robust low-frequency structure established by the initial generation while focusing its generative capacity on producing higher-fidelity high-frequency details. By guiding the initial stages of the refinement in this manner, we effectively reduce common visual artifacts, enhance sharpness, and improve the overall perceptual quality of the generated urban scenes without incurring any additional training costs.

This method increases sampling time but allows step adjustment. The standard denoising step is 50 with CFG at 100 Number of Function Evaluations~(NFEs). We use 30 steps without CFG for the first denoising pass and 30 steps with CFG for the second pass at 90 NFE, achieving 10\% NFE reduction while improving sampling quality.

While AAM demonstrates excellent performance in generating high-quality individual frames, making it particularly suitable for image-to-video (I2V) conversion and high-quality synthetic data production, it exhibits significant limitations in autoregressive long-video generation scenarios, often resulting in discontinuity between generated frames and historical frames. This limitation may stem from the fact that AAM's pretrained diffusion model is fundamentally based on an I2V architecture rather than a video-to-video (V2V) framework. Fine-tuning AAM's pretrained model on V2V tasks could potentially alleviate this issue.

\subsubsection{Time Travel Sampling based on SDE~(TTS-SDE) for Enhanced Video Generation}
Inspired by DDNM~\cite{wang2022zero}, Repaint~\cite{lugmayr2022repaint}, and OSV~\cite{mao2025osv} approaches, we introduced a novel high-quality sampling methodology Time Travel Sampling based on SDE~(TTS-SDE), as shown in Algorithm~\ref{alg:time_travel_sampling}. For a selected timestep $t_{n}$, we first sample forward $l$ steps to obtain $x_{t_{\text{max}(n-l,0)}}$, then leverage this ``future" state's more accurate velocity vector estimation to reconstruct $x_{t_{n-1}}$, effectively utilizing prospective information to refine past states.

However, given the deterministic nature of ODE-based sampling with fixed noise inputs, this method's capability to enhance textual controllability remains limited, primarily improving only the sharpness of generated videos. We find that replacing ODE with SDE sampling significantly boosts textual controllability by introducing controlled stochasticity into the generation process.

\subsection{Camera Motion Control}
\label{sec:qcmm}
Existing approaches for camera control in video generation often rely on providing a dense sequence of per-frame absolute camera-to-world (c2w) pose matrices~\cite{wang2023motionctrl, he2024cameractrl}. While offering explicit control, this representation can be overly granular, potentially leading to less stable or unintuitive camera trajectories, and may not effectively capture the inherent temporal coherence of continuous camera movements.

Instead of directly using continuous pose matrices, we introduced a discrete camera motion representation, as introduced in Section~\ref{sec:Method_Quantized_Camera_Motion}. We quantize the camera motion and parse the motion into text as a textual condition.

In addition, beyond the motion direction, we propose to control the motion speed to achieve stable movement in the generated videos. Specifically, as outlined in Algorithm~\ref{alg:cmq}, for each camera trajectory, we compute three quantitative indicators of the motion speed:

\textbf{Translational Motion} ($\mathcal{V}$) is captured by the displacement vector between the positions of two consecutive frames, which quantifies the local translational changes along the camera trajectory. By explicitly modeling this motion, the model learns to sustain consistent forward movements or lateral shifts, thereby preserving spatial continuity and fostering a natural flow in the generated videos.

\textbf{Directional Change} ($\mathcal{D}$) is measured by the angle formed between displacement vectors across every three consecutive frames, effectively measuring local turning or bending of the trajectory. Incorporating this measure encourages the model to produce smooth directional transitions and avoid abrupt changes, resulting in more realistic and stable motion paths.

\textbf{Rotational Dynamics} ($\mathcal{R}$) are characterized by the change in camera orientation angle between two consecutive frames, reflecting the camera’s rotational behavior. By explicitly incorporating this rotational cue, the model can better synchronize viewpoint adjustments with the underlying motion dynamics, leading to natural camera panning, tilting, and turning movements.

We also parse these speeds into the text, and they are explicitly provided to the model during training and fixed during inference to prevent the generation of videos with irregular or fluctuating speeds. We then combine it with the discrete camera motion description as the final camera motion condition. It effectively quantizes the continuous camera trajectory into a sequence of semantically meaningful actions, inherently integrating temporal context from the relative pose changes.

The camera motion condition is then combined with a prefix text description of the video (``This video depicts a city walk scene with a first-person view") as the final textual condition.

\begin{algorithm}
\caption{Camera Motion Speed Calculation}
\label{alg:cmq}
\begin{tabular}{rl}
\multicolumn{2}{l}{{\bf Require:} Sequence of camera-to-world matrices $\mathcal{C} = \{C_0, C_1, \ldots, C_{N-1}\}$.} \\
1: & Initialize $\mathcal{V} \leftarrow \emptyset$, $\mathcal{D} \leftarrow \emptyset$, $\mathcal{R} \leftarrow \emptyset$ \\
2: & {\bf for} $i = 0$ to $N - 3$ {\bf do} \\
3: & \quad $p_i \leftarrow$ $C_i[:, :3, 3]$ \\
4: & \quad $p_{i+1} \leftarrow$ $C_{i+1}[:, :3, 3]$ \\
5: & \quad $p_{i+2} \leftarrow$ $C_{i+2}[:, :3, 3]$ \\
6: & \quad $v_1 \leftarrow p_{i+1} - p_i$ \\
7: & \quad $v_2 \leftarrow p_{i+2} - p_{i+1}$ \\
8: & \quad Compute $\phi = \arccos \left(\frac{v_1 \cdot v_2}{|| v_1 || || v_2 ||}\right)$ \\
9: & \quad Append $v_1$ to $\mathcal{V}$ \quad // Translational vector change \\
10: & \quad Append $\phi$ to $\mathcal{D}$ \quad // Directional angle change \\
11: & {\bf end for} \\
12: & $p_{N-2} \leftarrow$ $C_{N-2}[:, :3, 3]$ \\
13: & $p_{N-1} \leftarrow$ $C_{N-1}[:, :3, 3]$ \\
14: & $v \leftarrow p_{N-1} - p_{N-2}$ \\
15: & Append $v$ to $\mathcal{V}$\\
16: & {\bf for} $i = 0$ to $N - 2$ {\bf do} \\
17: & \quad $z_i \leftarrow$ $C_i[:, 2]$\\
18: & \quad $z_{i+1} \leftarrow$ $C_{i+1}[:, 2]$ \\
19: & \quad Compute $\theta = \arccos \left(\frac{z_i \cdot z_{i+1}}{|| z_i || || z_{i+1} ||}\right)$ \\
20: & \quad Append $\theta$ to $\mathcal{R}$ \quad // Rotation angle change \\
21: & {\bf end for} \\
22: & {\bf return} $\mathcal{V}, \mathcal{D}, \mathcal{R}$
\end{tabular}
\end{algorithm}

\subsection{Application}
We show some applications in the demo video on the project page.

\subsubsection{World Generalization}
Though \method is trained by real-world videos, it shows impressive generalizability to diverse unreal scenes, such as animation, video games, and AI-generated images. Thus, \method allows not only real-world exploration, but also facilitates unreal-world exploration.
Moreover, \method supports V2V and thus also can adapt to live images taken by iPhone.

\subsubsection{World Editing}
Since \method shows strong generalizability and thus we can achieve world editing by simply combining \method with image editing methods such as GPT-4o. We show some examples that change weather, time, and style during video generation using GPT-4o in the demo video.


\subsection{Acceleration}
We present an acceleration framework for the \method model. Departing from conventional approaches, we devise a co-optimization strategy integrating step distillation with cache acceleration. This design stems from an insight: aggressively reducing sampling steps to just one step severely compromises the model's adaptability to diverse complex scenarios, while video quality progressively deteriorates due to error accumulation as generation length increases. To address this, we reduce sampling steps to 14 while incorporating a cache acceleration mechanism.

\subsubsection{Adversarial Distillation for Accelerated Diffusion Sampling}
\label{sec:gan_distillation}

We introduce an adversarial distillation framework to reduce the sampling steps while preserving visual quality. The core innovation leverages Generative Adversarial Networks (GANs) to distill the iterative denoising process into fewer steps, following the formulation:

\begin{equation}
\mathcal{L}_{\text{total}} = \mathcal{L}_{\text{diffusion}} + \lambda_{\text{adv}}\mathcal{L}_{\text{adv}}
\end{equation}

where $\mathcal{L}_{\text{diffusion}}$ is the standard diffusion loss and $\mathcal{L}_{\text{adv}}$ is the adversarial loss term. The training alternates between updating the denoiser DiT Model and the discriminator $\mathcal{D}$.

The discriminator $\mathcal{D}$ is trained to distinguish between real samples ${x}_{\text{real}}$ from the training distribution and denoised samples $\hat{{x}}_0$ generated by the diffusion model:

\begin{align}
\hat{{x}}_0 &= \mathbf{z}_t - t \cdot v_{\theta}(\mathbf{z}_t, t, \mathbf{c}) 
\end{align}

The discriminator loss combines feature-level and image-level discrimination:
\begin{equation}
\mathcal{L}_{\mathcal{D}} = \mathbb{E}[\text{ReLU}(1 - \mathcal{D}({x}_{\text{real}}))] + \mathbb{E}[\text{ReLU}(1 + \mathcal{D}(\hat{{x}}_0))]
\end{equation}

where $\mathcal{D}$ returns both image-level predictions and intermediate feature maps.

The denoiser DiT Model \method aims to fool the discriminator while maintaining denoising accuracy:

\begin{equation}
\mathcal{L}_{\text{adv}} = -\mathbb{E}[\mathcal{D}(\hat{{x}}_0)]
\end{equation}

We adopt the discriminator design from OSV~\cite{mao2025osv}, as this approach significantly reduces memory consumption while achieving excellent discriminative performance.

\subsubsection{Cache-Accelerating}
We use the caching mechanism that reduces computational redundancy by reusing intermediate residual features across denoising steps. The system employs layer-specific caching policies to achieve computation reduction while preserving output quality.

For layer  $l$  at timestep  $t_n$ , the residual feature  $\Delta{x}^l_{t_n}$ is cached when: 
\begin{equation}
\mathcal{C}^l_{t_n} = \begin{cases} 
(\text{Block}^l({x}^{l-1}_{t_n}) - {x}^{l-1}_{t_n})_{\text{bfloat16}} & \text{if } l \in \mathcal{L}_{\text{cache}} \\
\emptyset & \text{otherwise}
\end{cases}
\end{equation}
where $\mathcal{L}_{\text{cache}}$ denotes predefined cacheable layers. These cached residuals are stored in $\texttt{bfloat16}$ precision.

\begin{figure*}[htp]
    \centering
    \includegraphics[width=1.0\linewidth]{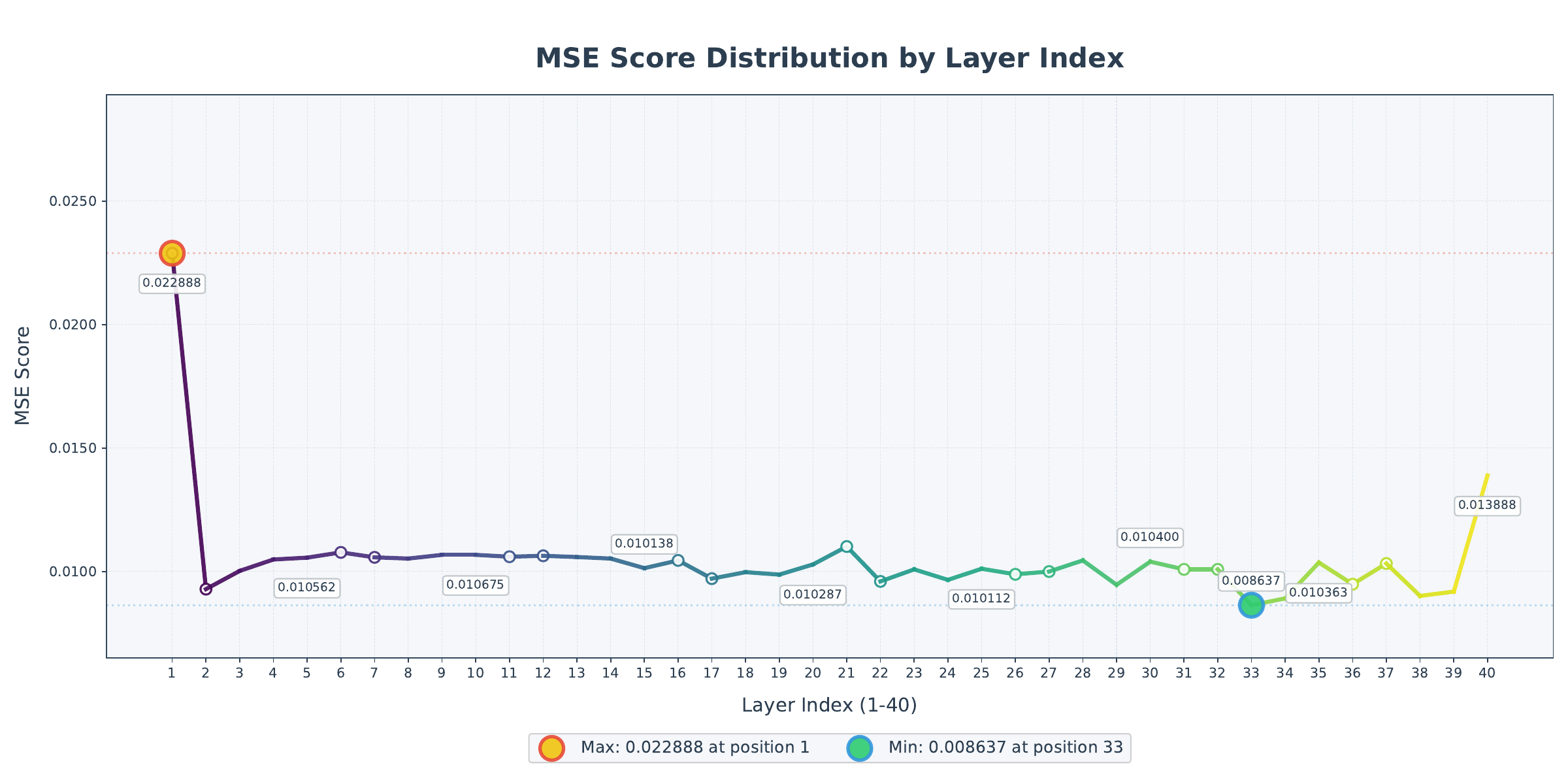}
    \caption{Significance of Individual DiT Blocks.}
    \label{fig:cache_index}
\end{figure*}

At subsequent timestep $t_{n-1}$, layer $l$ computes:  
\begin{equation}
{x}^l_{t_{n-1}} = \begin{cases} 
{x}^{l-1}_{t_{n-1}} + \mathcal{C}^l_{t_n} & \text{when } l \in \mathcal{L}_{\text{cache}} \\
\text{Block}^l({x}^{l-1}_{t_{n-1}}) & \text{otherwise}
\end{cases}
\end{equation}

We consider an acceleration ratio of $1:l_c$, whereby full computation is performed at time step $t_n$ to obtain cache $\mathcal{C}^l_{t_n}$, after which computations for subsequent steps $t_{n-1}$ through $t_{n-l_c}$ utilize $\mathcal{C}^l_{t_n}$ for skip-step processing. To evaluate the impact of individual blocks within the DiT denoising model, we introduce an MSE-based importance metric. Specifically, across $N/(l_c+1)$ cached computation segments from $t_n$ to $t_{n-l_c}$, upon computing the reference cache $\mathcal{C}^l_{t_n}$ at $t_n$, we systematically ablate each of the 40 DiT blocks at step $t_{n-l_c}$. The influence score for the $i$-th block is quantified as:  
\begin{equation}  
\text{MSE Score}_i = \text{Mean} \left( \left| {x}^{\text{Remove } i}_{t_{n-l_c}} - {x}_{t_{n-l_c}} \right|^{2}_{2} \right), i \in [1,2,3,..,40]
\end{equation}
where Mean denotes the averaging operation. Each video yields $N/(l_c+1)$ measurements per block, generating 40 temporal MSE profiles per segment. These profiles are averaged across time steps and aggregated over 32 videos to produce 40 composite MSE scores. Figure~\textcolor{red}{\ref{fig:cache_index}} illustrates that central DiT blocks exhibit minimal impact while initial and terminal blocks demonstrate maximal influence, leading to the selection of the 10 lowest-scoring blocks as predefined cacheable layers.

\section{Experiment}
\label{sec:Experiment}

\subsection{Experimental Settings}
\subsubsection{Training Details}
We utilized the SkyReels-V2-14B-540P as the pre-trained model.
The training process involved video resolutions of 544$\times$960, with a frame rate of 16 FPS, a batch size of 40, and the Adam optimizer with a learning rate of 1e-5. 
Training was conducted across NVIDIA A800 GPUs over 7,000 iterations.
\subsubsection{Evaluation Dataset}
Existing video generation evaluation methods are not well-suited for complex scenes and interactive generation with keyboard inputs. To address this issue, we developed the Yume-Bench evaluation framework. Specifically, we exclude training videos from the Sekai-Real-HQ dataset and instead sample test videos with quantized camera motions. For rare actions, such as walking backward or tilting the camera up and down, we randomly sample images rather than video clips. In total, we collected 70 videos or images, covering a wide range of complex action combinations, as detailed in Table~\textcolor{red}{\ref{tab:action_combinations}}. 
\subsubsection{Evaluation Details}
Yume-Bench evaluates two core capabilities of models: visual quality and instruction following (camera motion tracking), using six fine-grained metrics.
In the instruction following evaluation, we assess whether the generated videos correctly follow the intended walking direction and camera movements. While camera poses estimation methods, such as MegaSaM~\cite{li2025megasam}, can automate this evaluation, the camera motion prediction in the generated videos is not sufficiently accurate, and quantization errors may occur. Therefore, we conduct a human evaluation to identify the accuracy of the generated motion.
For the remaining metrics, we adopt those from VBench~\cite{huang2024vbench}, including subject consistency, background consistency, motion smoothness, aesthetic quality, and imaging quality.
The test data resolution is 544$\times$960 with a frame rate of 16 FPS, comprising a total of 96 frames.
We applied 50 inference steps for all models tested.

\begin{table}[h]
\centering
\caption{Keyboard-Mouse Action Combinations}
\label{tab:action_combinations}
\begin{tabular}{lc}
\toprule
\textbf{Keyboard-Mouse Action} & \textbf{Count} \\ 
\midrule
No Keys + Mouse Down & 2 \\
No Keys + Mouse Up & 2 \\
S Key + No Mouse Movement & 2 \\
W+A Keys + No Mouse Movement & 29 \\
W+A Keys + Mouse Left & 6 \\
W+A Keys + Mouse Right & 17 \\
W+D Keys + No Mouse Movement & 5 \\
W+D Keys + Mouse Left & 5 \\
W+D Keys + Mouse Right & 2 \\
\bottomrule
\end{tabular}
\end{table}

\subsection{Qualitative Results}
\subsubsection{Image-to-Video Generation}
We compared several state-of-the-art (SOTA) image-to-video generation models, including Wan-2.1 and MatrixGame, as shown in Table \ref{tab:1}.
Our experimental results revealed that:
(1) Wan-2.1 shows limited instruction-following capabilities while using textual instructions to control camera motion. 
(2) Although MatrixGame demonstrates some degree of controllability, it struggles to generalize to real-world scenarios and lacks sufficient scene replication control. In contrast, our \method excels in controllability, with its instruction-following capability scoring 0.657, significantly outperforming other models.
Additionally, \method achieves optimal or near-optimal performance across other metrics, demonstrating our superior visual quality.

\begin{figure*}[htp]
    \centering
    \includegraphics[width=1.0\linewidth]{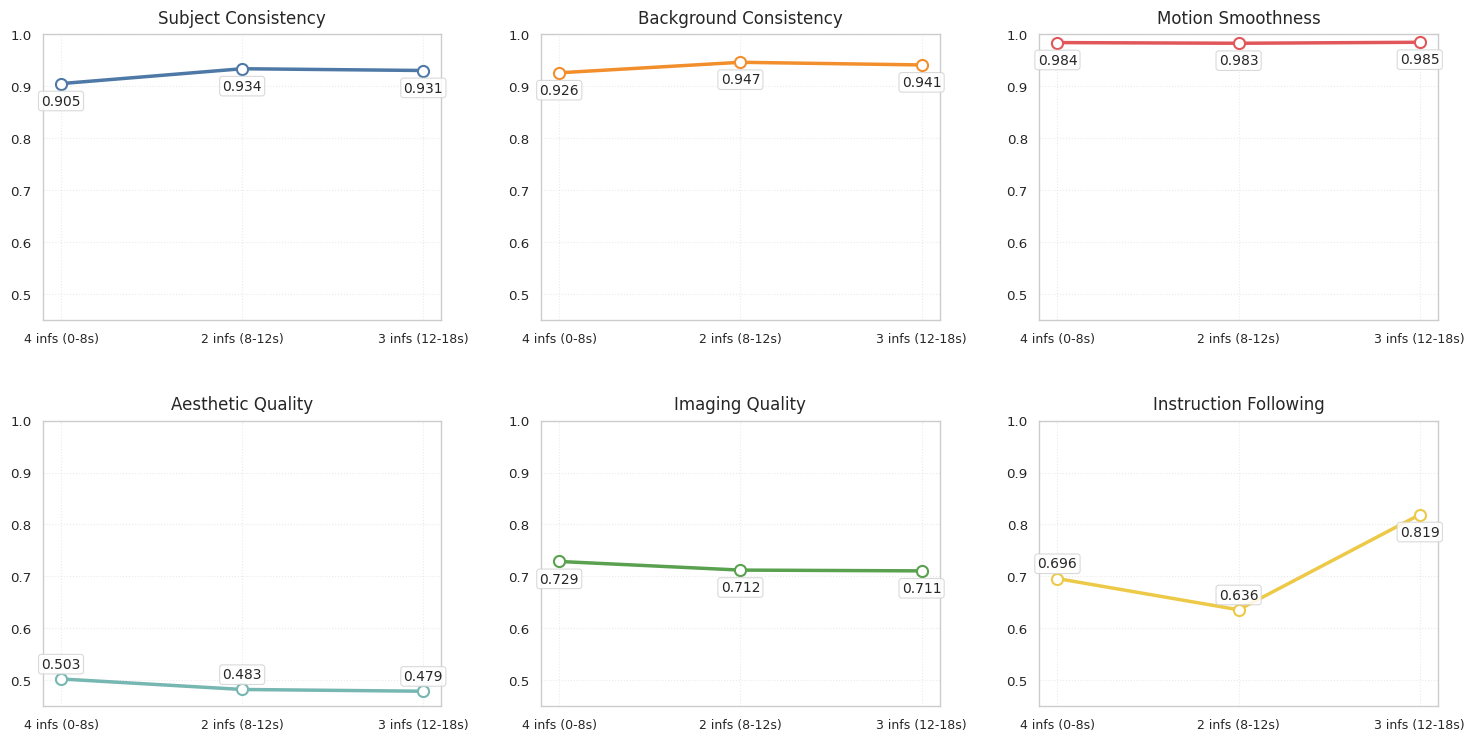}
    \caption{Metric Dynamics in Long-video Generation. We use TTS-SDE. We performed a total of 9 extrapolations. "4 infs" represents using videos obtained from 4 extrapolations (totaling 8 seconds) for metric calculation, while "2 infs" represents using videos obtained from 2 extrapolations (also totaling 4 seconds) for metric calculation.}
    \label{fig:long_video}
\end{figure*}


\begin{table}[]
  \caption{Quality comparison of different models. Wan-2.1 utilize text-based control. MatrixGame employs its own native keyboard/mouse control scheme. All models use the same random seed.}
  \label{tab:1}
  \centering
  \resizebox{\linewidth}{!}{
  \begin{tabular}{ccccccc}
    \toprule
    \multirow{1}{*}{Model}
    & \makecell{Instruction\\Following $\uparrow$} & \makecell{Subject\\Consistency $\uparrow$} & \makecell{Background\\Consistency $\uparrow$} & \makecell{Motion\\Smoothness $\uparrow$} & \makecell{Aesthetic\\Quality $\uparrow$} & \makecell{Imaging\\Quality $\uparrow$} \\
    \midrule
    Wan-2.1\multirow{1}*{~\cite{wan2025wan}} &0.057 &0.859 &0.899 &0.961 &0.494 &0.695\\
    \midrule
MatrixGame~\cite{zhang2025matrixgame} &0.271 &0.911 &0.932 &0.983 &0.435 &\textbf{0.750} \\
    \midrule
    \method~(Ours)~&\textbf{0.657}&\textbf{0.932} &\textbf{0.941} &\textbf{0.986} &\textbf{0.518} &0.739 \\
    \bottomrule
  \end{tabular}}
\end{table}

\subsubsection{Validation of Long-video Generation Performance}

To assess the long-video generation capability, we created an 18-second video sequence where \method generates 2-second segments incrementally.
During the first 8 seconds, the motion patterns remained consistent with the test set, followed by a transition to continuous forward movement (W) in the subsequent 10 seconds. As shown in Figure~\textcolor{red}{\ref{fig:long_video}}, mild content decay was observed: subject consistency decreased by 0.5\% (0.934→0.930), and background consistency dropped by 0.6\% (0.947→0.941) between the 0-8s and 12-18s segments, indicating that \method maintains reasonable stability over time.
It is worth noting that during the motion transition phase (8-12s), instruction-following performance dropped by 8.6\% (0.947→0.941). This decline can be attributed to the inertia from the motion in the input video, which hindered an immediate reversal of direction. However, the inertial effect diminished after 12 seconds, leading to a significant recovery in instruction-following performance by 22.3\% (0.636→0.819).

\begin{table}[htp]
  \caption{Ablation study on different samplers.}
  \label{tab:model_comparison}
  \centering
  \resizebox{\linewidth}{!}{
  \begin{tabular}{lccccccc}
    \toprule
    \multirow{1}{*}{Model}
    & \makecell{Instruction\\Following $\uparrow$} & \makecell{Subject\\Consistency $\uparrow$} & \makecell{Background\\Consistency $\uparrow$} & \makecell{Motion\\Smoothness $\uparrow$} & \makecell{Aesthetic\\Quality $\uparrow$} & \makecell{Imaging\\Quality $\uparrow$} \\
    \midrule
    \method-ODE & 0.657 & \textbf{0.932} & \textbf{0.941} & \textbf{0.986} & 0.518 & \textbf{0.739} \\
    \method-SDE & 0.629 & 0.927 & 0.938 & 0.985 & 0.516 & 0.737 \\
    \method-TTS-ODE & 0.671 & 0.923 & 0.936 & 0.985 & \textbf{0.521} & 0.737 \\
    \method-TTS-SDE & \textbf{0.743} & 0.921 & 0.933 & 0.985 & 0.507 & 0.732 \\
    \bottomrule
  \end{tabular}}
\end{table}

\begin{figure*}[h]
    \centering
    \includegraphics[width=1.0\linewidth]{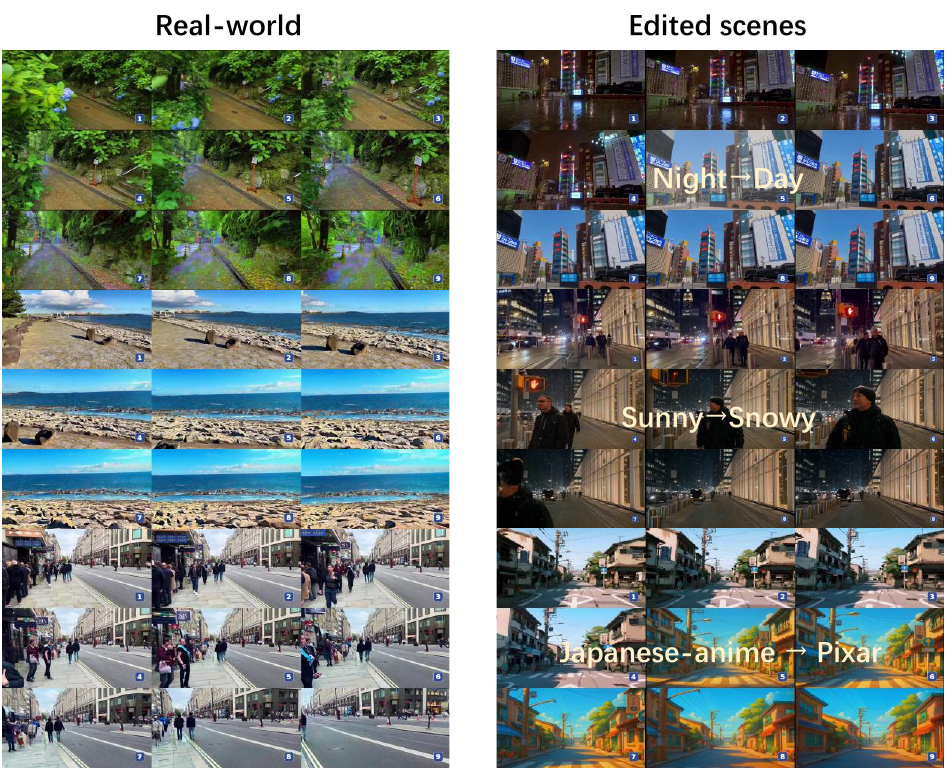}    \caption{\method demonstrates superior visual quality and precise adherence to keyboard control in real-world and unreal scenarios.}
    \label{fig:vis}
\end{figure*}



\subsection{Ablation study}
\subsubsection{Verification of TTS-SDE Effectiveness}
To assess the effectiveness of TTS-SDE, we replaced the ODE sampling with SDE and TTS-SDE for comparison. As shown in Table~\textcolor{red}{\ref{tab:model_comparison}}, while SDE sampling resulted in a decline across all metrics, TTS-SDE achieved a significant improvement in instruction following, despite a slight reduction in other indicators. This indicates that TTS-SDE strategically introduces noise perturbations, enhancing the refinement of motion trajectories in the generated videos. Furthermore, the integration of TTS-SDE has resulted in improved aesthetic scores, with our observations revealing clearer and more detailed generated videos.

\begin{table}[htp]
  \caption{Validation of distillation method effectiveness.}
  \label{tab:model_comparison}
  \centering
  \resizebox{\linewidth}{!}{
  \begin{tabular}{lccccccc}
    \toprule
    Model & 
    \makecell{Time~(s)$\downarrow$} & 
    \makecell{Instruction\\Following $\uparrow$} & 
    \makecell{Subject\\Consistency $\uparrow$} & 
    \makecell{Background\\Consistency $\uparrow$} & 
    \makecell{Motion\\Smoothness $\uparrow$} & 
    \makecell{Aesthetic\\Quality $\uparrow$} & 
    \makecell{Imaging\\Quality $\uparrow$} \\
    \midrule
    {Baseline} &583.1 &\textbf{0.657} & \textbf{0.932} & \textbf{0.941} & \textbf{0.986} & 0.518 & \textbf{0.739} \\
    {Distil} &\textbf{158.8} &0.557 & 0.927 & 0.940 & 0.984 & \textbf{0.519} &\textbf{0.739} \\
    \bottomrule
  \end{tabular}}
\end{table}

\subsubsection{Validating the effect of model distillation}
After distilling the model to reduce the number of steps from 50 to 14, we compared it with the original model. We found that, except for instruction following, the other metrics showed minimal differences from the original model. This may be because fewer steps weaken the model's text-control capability. 


\subsection{Visualization Results}
\begin{figure}[h]
    \centering
    \includegraphics[width=0.7\linewidth]{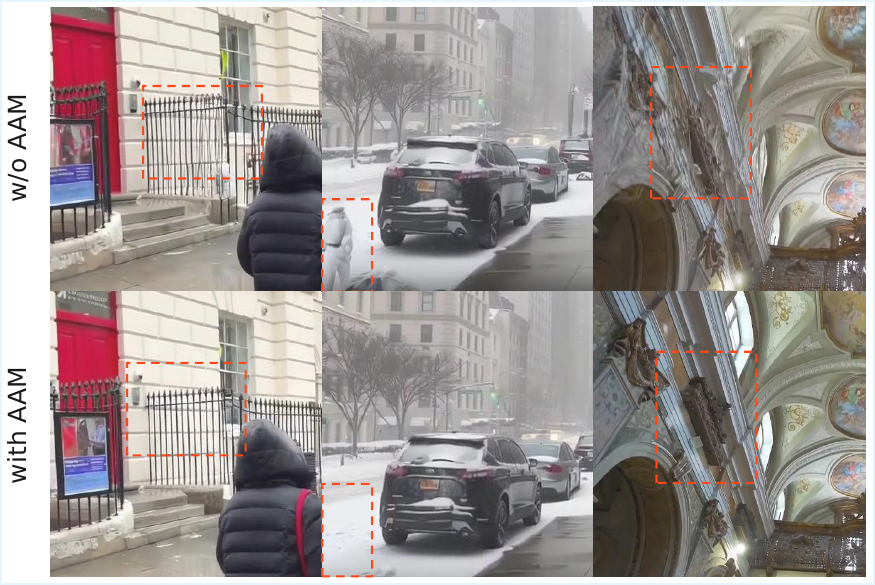}
    \caption{AAM Improves Structural Details in Urban and Architectural Scenes.}
    \label{fig:vis2}
\end{figure}

As illustrated in Figure\textcolor{red}{~\ref{fig:vis}}, we generated multiple video sequences using the initial frame image and quantized camera trajectories, demonstrating that \method accurately follows the predefined motion paths during video generation. Figure\textcolor{red}{~\ref{fig:vis2}} demonstrates the effectiveness of AAM by generating clearer videos while avoiding illogical scenes such as aberrant snowman artifacts.

\section{Conclusion}
\label{sec:conclusion}
In this paper, we introduce a preview version of \method, which is an interactive world generation model that allows the use of keyboard inputs to explore a dynamic world created by an input image. Moreover, it can do infinite video generation in an autoregressive manner. \method consists of four main components, including camera motion quantization, video generation architecture, advanced sampler, and model acceleration. 

\method is a long-term project that has established a solid foundation, yet still faces numerous challenges to address, such as the visual quality, runtime efficiency, and control accuracy. Moreover, many functions need to be achieved, such as interaction with objects.

\clearpage

\appendix

\definecolor{pytorchblue}{RGB}{39, 89, 191}
\definecolor{pytorchgreen}{RGB}{0, 128, 0}
\definecolor{pytorchred}{RGB}{200, 0, 0}

\lstdefinestyle{pytorch}{
    language=Python,
    basicstyle=\ttfamily\small,
    keywordstyle=\color{pytorchblue},
    commentstyle=\color{pytorchgreen},
    stringstyle=\color{pytorchred},
    showstringspaces=false,
    breaklines=true,
    frame=lines,
    numbers=left,
    stepnumber=1,
    numbersep=5pt,
    tabsize=4,
    morekeywords={torch, Tensor, matmul, movedim, linalg, svd, where, zeros, diag, device, cuda, is_available}
}

\clearpage

\section{Ablation Study on Image to Video}
\begin{table}[htp]
  \caption{Effect of Controlled Condition Injection Methods.}
  \label{tab:2}
  \centering
  \resizebox{\linewidth}{!}{
  \begin{tabular}{cccccc}
    \toprule
      Name  & instruction following $\uparrow$ &aesthetic quality $\uparrow$ &imaging quality $\uparrow$ &motion smoothness $\uparrow$ &background consistency $\uparrow$\\
    \midrule
    AdaLN-Zero & 0.35 &\textbf{0.517} & 0.694 &0.988 &\textbf{0.936} \\
    \midrule
    Cross-Attention &\textbf{0.45} &0.507 &\textbf{0.701} & 0.987 &0.935 \\
    \midrule
    Text injection &\textbf{0.45} &0.492 &0.694 &\textbf{0.991} &0.902 \\
    \bottomrule
  \end{tabular}}
\end{table}
We validated the effectiveness of the MVDT and AAM modules within our image-to-video (I2V) pipeline through comparative analysis of 20 randomly selected video sequences. For this experiment, all models were trained exclusively for 1,000 iterations.

\noindent\textbf{Effect of Controlled Condition Injection Methods. }We replace \method's MVDT architecture with DiT and removed the text injection approach as Baseline-1. To evaluate different controlled condition injection methods, we incorporated adaLN-zero, cross-attention, and text injection into Baseline-1 while maintaining identical training parameters. As shown in Table\textcolor{red}{~\ref{tab:2}}, these methods demonstrated complementary advantages in V-bench metrics. We adopt the text injection approach due to its superior controllability, seamless integration with pretrained models (requiring no architectural modifications), and parameter-efficient design that introduces no additional learnable parameters.

\begin{table}[htp]
  \caption{Effect of MVDT.}
  \label{tab:3}
  \centering
  \resizebox{\linewidth}{!}{
  \begin{tabular}{ccccc}
    \toprule
      MVDT  $\downarrow$ &aesthetic quality $\uparrow$ &imaging quality $\uparrow$ &motion smoothness $\uparrow$ &background consistency $\uparrow$\\
    \midrule
    + &\textbf{0.517} &\textbf{0.702} &0.985 &\textbf{0.929} \\
    \midrule
    -  &0.492 &0.694 &\textbf{0.991} &0.902 \\
    \bottomrule
  \end{tabular}}
\end{table}

\noindent\textbf{Effect of MVDT Architecture. }We validated the efficacy of MVDT by comparing \method performance with and without this architecture in identical training configurations, we deliberately excluded the AAM in all experimental configurations. Table\textcolor{red}{~\ref{tab:3}} confirms that the MVDT structure consistently enhances \method's Generate capabilities.  The MVDT architecture enhances \method's ability to capture structural relationships between frames, significantly reducing artifacts in generated videos.

\begin{table}[htp]
  \caption{Effect of Anti-Artifact Mechanism~(AAM).}
  \label{tab:4}
  \centering
  \resizebox{\linewidth}{!}{
  \begin{tabular}{cccccc}
    \toprule
      AAM   $\uparrow$ &aesthetic quality $\uparrow$ &imaging quality $\uparrow$ &motion smoothness $\uparrow$ &dynamic degree $\uparrow$\\
    \midrule
    + &\textbf{0.529} &\textbf{0.737} &\textbf{0.986} &\textbf{0.937}  \\
    \midrule
    - &0.517 &0.702 &0.985 &0.929 \\
    \bottomrule
  \end{tabular}}
\end{table}

\begin{figure}[htp]
    \centering
    \includegraphics[width=0.7\linewidth]{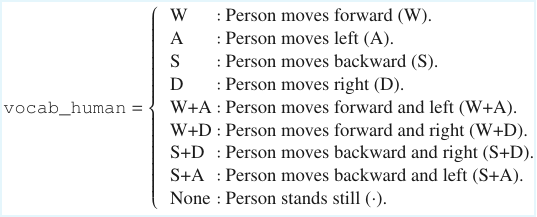}
    \caption{vocab example for translational motion.}
    \label{fig:vocab_human}
\end{figure}

\begin{figure}[htp]
    \centering
    \includegraphics[width=0.7\linewidth]{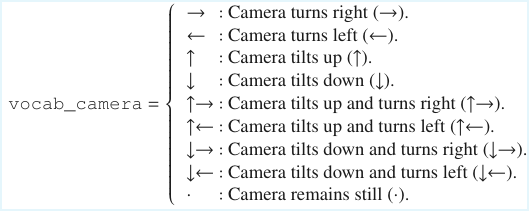}
    \caption{vocab example for rotational motion.}
    \label{fig:vocab_camera}
\end{figure}

\section{Quantized Camera Motion of Text}
We generate natural language descriptions for human movement and camera motion like the example in \texttt{vocab\_human} and \texttt{vocab\_camera} of Figure ~\ref{fig:vocab_human}, ~\ref{fig:vocab_camera}.

\section{Action Distribution Statistics}
\begin{figure*}[!h]
    \centering
    \includegraphics[width=\linewidth]{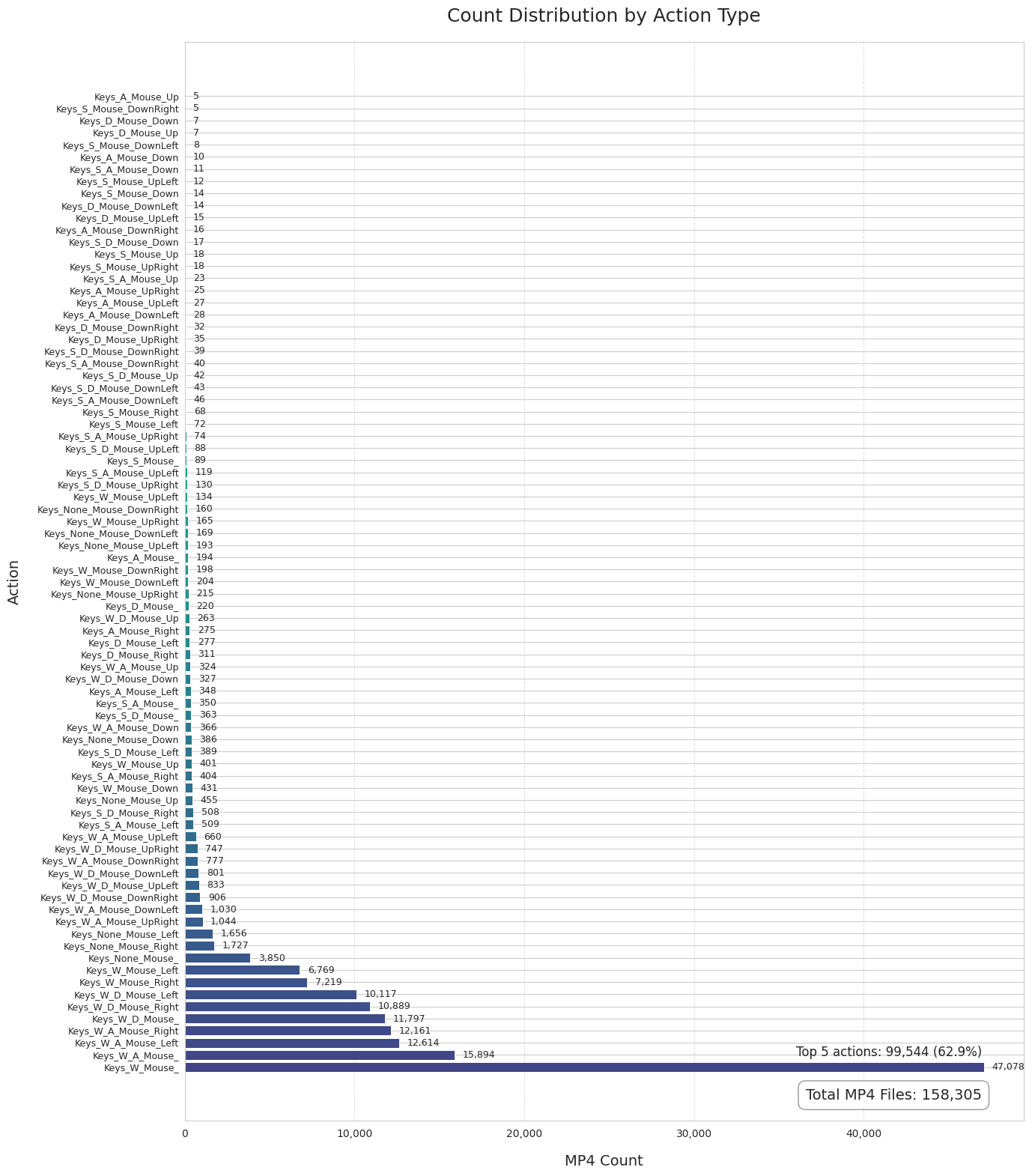}
    \caption{Overview of action distribution.}
    \label{fig:distribution}
\end{figure*}

Figure~\textcolor{red}{\ref{fig:distribution}} presents the distribution of 21,526 video clips across different action combinations.

\section{Core Algorithm Implementation of Gaussian Blur Kernel}
We implement a separable 2D linear operator using height/width blur kernels with SVD decomposition. The projection $B(z) = A^{\text{Pinv}}Az$ extracts low-frequency components from first-stage results, while the null-space projection $z-B(z) = (I- A^{\text{Pinv}}A)z$ preserves high-frequency details in second-stage outputs. We provide PyTorch-style implementation code.

\begin{lstlisting}[style=pytorch]
def project_null_space(x: torch.Tensor, 
                      A_operator: 'LinearOperator2D') -> torch.Tensor:

    original_shape = x.shape
    x_flat = x.view(-1, *original_shape[-2:])
    

    Ax = A_operator.A(x_flat)
    A_Pinv_Ax = A_operator.A_inv(Ax)
    

    I_A_Pinv_A_x = x_flat - A_Pinv_Ax
    return I_A_Pinv_A_x.view(original_shape)
\end{lstlisting}

\section*{LinearOperator2D Class}

\begin{lstlisting}[style=pytorch]
class LinearOperator2D:
    def __init__(self, 
                 kernel_H: torch.Tensor,
                 kernel_W: torch.Tensor,
                 H: int, 
                 W: int,
                 device: str = None):
        """Initialize separable 2D operator with SVD decomposition"""
        self.device = device or ('cuda' if torch.cuda.is_available() else 'cpu')
        self.H, self.W = H, W
        
        # Height-direction operator A_H (H x H)
        A_H = torch.zeros(H, H, device=self.device)
        for i in range(H):
            for j in range(i - len(kernel_H)//2, i + len(kernel_H)//2 + 1):
                if 0 <= j < H:
                    A_H[i,j] = kernel_H[j - i + len(kernel_H)//2]
        
        U_H, S_H, Vt_H = torch.linalg.svd(A_H, full_matrices=False)
        self.U_H, self.S_H, self.Vt_H = U_H, S_H, Vt_H
        self.S_pinv_H = torch.where(S_H > 1e-6, 1/S_H, torch.zeros_like(S_H))
        
        # Width-direction operator A_W (W x W)
        A_W = torch.zeros(W, W, device=self.device)
        for i in range(W):
            for j in range(i - len(kernel_W)//2, i + len(kernel_W)//2 + 1):
                if 0 <= j < W:
                    A_W[i,j] = kernel_W[j - i + len(kernel_W)//2]
        
        U_W, S_W, Vt_W = torch.linalg.svd(A_W, full_matrices=False)
        self.U_W, self.S_W, self.Vt_W = U_W, S_W, Vt_W
        self.S_pinv_W = torch.where(S_W > 1e-6, 1/S_W, torch.zeros_like(S_W))
\end{lstlisting}

\section*{Operator Application Methods}

\begin{lstlisting}[style=pytorch]
    def A(self, x: torch.Tensor) -> torch.Tensor:
        """Forward operation: A = A_W @ A_H"""
        # Height processing
        x_h = x.movedim(-2, -1)  # [..., W, H]
        x_h = torch.matmul(x_h, self.Vt_H.T)
        x_h = torch.matmul(x_h, torch.diag(self.S_H))
        x_h = torch.matmul(x_h, self.U_H.T)
        x_h = x_h.movedim(-1, -2)  # [..., H, W]
        
        # Width processing
        x_hw = torch.matmul(x_h, self.Vt_W.T)
        x_hw = torch.matmul(x_hw, torch.diag(self.S_W))
        x_hw = torch.matmul(x_hw, self.U_W.T)
        return x_hw

    def A_inv(self, y: torch.Tensor) -> torch.Tensor:
        """Pseudoinverse operation"""
        # Width pseudoinverse
        y_w = torch.matmul(y, self.U_W)
        y_w = torch.matmul(y_w, torch.diag(self.S_pinv_W))
        y_w = torch.matmul(y_w, self.Vt_W)
        
        # Height pseudoinverse
        y_hw = y_w.movedim(-2, -1)  # [..., W, H]
        y_hw = torch.matmul(y_hw, self.U_H)
        y_hw = torch.matmul(y_hw, torch.diag(self.S_pinv_H))
        y_hw = torch.matmul(y_hw, self.Vt_H)
        return y_hw.movedim(-1, -2)  # [..., H, W]
\end{lstlisting}

\section*{Operator Initialization Example}

\begin{lstlisting}[style=pytorch]
# Blur kernels
kernel_H = torch.tensor([0.1, 0.8, 0.1], device='cpu')  # Height kernel
kernel_W = torch.tensor([0.2, 0.6, 0.2], device='cpu')  # Width kernel

# Operator instantiation
A_op = LinearOperator2D(
    kernel_H=kernel_H,
    kernel_W=kernel_W,
    H=544,  # Height dimension
    W=960,  # Width dimension
    device='cuda' if torch.cuda.is_available() else 'cpu'
)
\end{lstlisting}
 

\begin{figure*}[htp]
    \centering
    \includegraphics[width=0.9\linewidth]{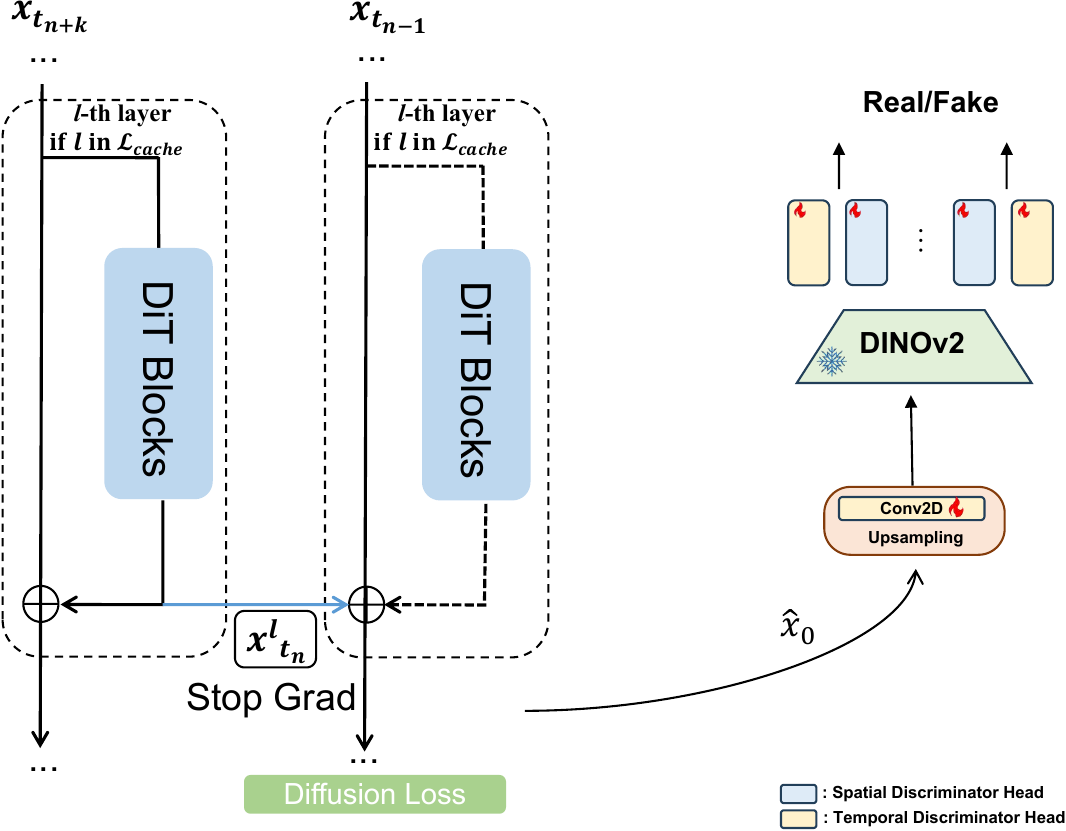}
    \caption{Acceleration Method Design.}
    \label{fig:Accelerated}
\end{figure*}

\section{The Joint Optimization of Adversarial Distillation and Caching}
We consider the joint optimization of adversarial distillation and caching. As illustrated in Figure~\textcolor{red}{\ref{fig:Accelerated}}, the DiT module performs intermediate feature caching for specific layers $l \in \mathcal{L}_{\text{cache}}$ during the state transition from $x_{t_{n+k}}$ to $x_{t_{n-1}}$ (where $t_{n+k} < 1$). The "Stop Grad" operation denotes gradient truncation, simulating scenarios where the DiT model encounters cached features during inference. This learning approach reduces errors when utilizing cached features. We optimize denoising fidelity using the Diffusion Loss while additionally incorporating the adversarial loss.

\newpage

\clearpage

\end{document}